\documentclass[journal]{IEEEtran}
\usepackage{amsmath,amsfonts}
\usepackage{algorithm}
\usepackage{array}
\usepackage[caption=false,font=normalsize,labelfont=sf,textfont=sf]{subfig}
\usepackage{textcomp}
\usepackage{stfloats}
\usepackage{url}
\usepackage{verbatim}
\usepackage{graphicx}
\usepackage{cite}
\usepackage[table]{xcolor}

\usepackage{amssymb}
\usepackage[noend]{algpseudocode}
\usepackage{multirow}
\usepackage[switch]{lineno}
\usepackage{color}
\usepackage{pifont}       
\usepackage{svg}

\graphicspath{{./imgs/}{./figures/}}

\newcommand{\revise}[1]{#1}

\begin{document}

\title{Tuning-Free Latent Diffusion Models for Ultrahigh-Resolution \\ Image Editing}

\author{Wanglong Lu, Lingming Su, Kaijie Shi,  Minglun Gong, Xiaogang Jin, Hanli Zhao$^{*}$, and Xianta Jiang
\thanks{This work was supported by the Zhejiang Provincial Natural Science Foundation of China (Grant No. LMS26F020042). X. Jin was supported by the National Natural Science Foundation of China (Grant No. U25A20440). X. Jiang was supported by the Natural Sciences and Engineering Research Council of Canada (NSERC, Grant No. DGECR-2020-00296).}
 \thanks{W. Lu is with the College of Computer Science and Artificial Intelligence, Wenzhou University, Wenzhou 325035, China, and also with the Department of Computer Science, Memorial University of Newfoundland, St. John's, NL A1B 3X5, Canada, and the AI Analytics Team, Nasdaq, St. John's, NL A1A 0L9, Canada.}
 \thanks{L. Su and H. Zhao are with the College of Computer Science and Artificial Intelligence, Wenzhou University, Wenzhou 325035, China.}
 \thanks{K. Shi and X. Jiang are with the Department of Computer Science, Memorial University of Newfoundland, St. John's, NL A1B 3X5, Canada.}
 \thanks{M. Gong is with the School of Computer Science University of Guelph Guelph, ON, N1G 2W1, Canada.}
 \thanks{X. Jin is with the State Key Laboratory of CAD\&CG, Zhejiang University, Hangzhou 310058, China.}
 \thanks{$^{*}$ Corresponding author. E-mail: hanlizhao@wzu.edu.cn}}

\maketitle

\begin{abstract}
Recent diffusion-based generative models have shown impressive performance in image generation and editing.
However, due to memory limitations and the high cost of collecting high-resolution training images, existing methods {are typically restricted to inputs with linear resolutions below 1K. In contrast, photos captured by modern mobile devices often reach linear resolutions up to 8K, revealing a significant gap between current capabilities and real-world demands.}
Simply upscaling low-resolution edited results often results in visually enlarged but blurry images that lack fine details.
This paper introduces UltraDiffEdit, a novel, tuning-free image editing framework that extends off-the-shelf latent diffusion models (LDMs) to ultra-high resolutions.
UltraDiffEdit employs {a multi-scale progressive editing strategy}, iteratively blending high-resolution edited content with unedited areas in a coarse-to-fine manner.
We employ multi-patch encoding to preserve both edited and unedited visual details within the latent space.
To mitigate editing artifacts, our global-local consistency denoising technique consistently integrates edited and unedited latent features,
ensuring smooth transition at editing boundaries from {the latent representation to the final image.
We also introduce a patch-based hybrid sampling approach that captures} local, intermediate, and global features, ensuring semantic coherence and enhancing fine detail during denoising.
We conduct extensive experiments demonstrating UltraDiffEdit’s superior editing quality and flexibility:
{it can handle image resolutions up to 8K using only a single} NVIDIA GeForce RTX 3090 GPU. The source code is publicly available at \noindent\textcolor{blue}{\url{https://github.com/LonglongaaaGo/UltraDiffEdit}}.
\end{abstract}


\section{Introduction}\label{sec:intro}

\begin{figure}[t]
	\includegraphics[width=0.485\textwidth]{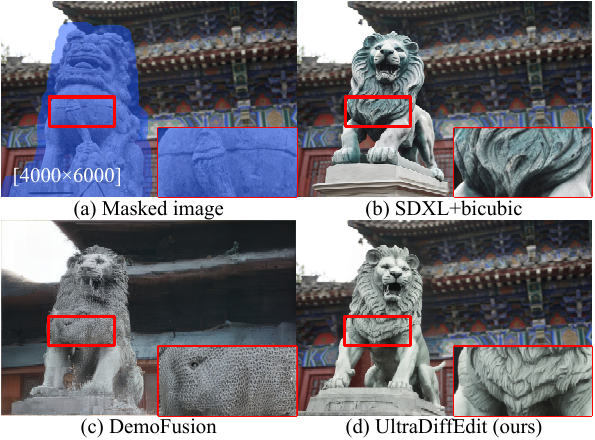}
     \caption{
    Examples of ultra-high-resolution image editing using SDXL~\cite{podell2023sdxl}:
    (a) An input image with masked regions (blue); 
    (b) This image is downscaled and processed with SDXL to yield a 1K edited result. Due to out-of-memory issues in SOTA super-resolution models (e.g., BSRGAN~\cite{zhang2021designing} and Inf-DiT~\cite{InfDiT}), the edited image is upscaled using the bicubic algorithm and merged with the original unmasked areas, resulting in limited fine detail. (c) DemoFusion~\cite{du2023demofusion} generates high-resolution images but struggles to preserve unedited parts, resulting in distorted details. (d) Our UltraDiffEdit provides seamless semantic editing, ensuring global-local consistency in real-world image editing.
    } 
	\label{fig:fig_show_demo}
\end{figure}

\IEEEPARstart{D}{iffusion}-based generative models~\cite{sohl2015deep, ho2020denoising} have been demonstrating remarkable capabilities in synthesizing high-quality and diverse images, opening up many possibilities for downstream applications such as image editing~\cite{zhang2023adding,Weather_TNNLS,Zhu_TNNLS,Liu_TNNLS,Wang_TNNLS,Pang_TNNLS}. 
Despite recent advances in diffusion models (e.g., LDMs~\cite{rombach2022high}), their application to high-resolution image editing~\cite{Kawar_2023_CVPR} {remains challenging}.
This is primarily because {most pre-trained diffusion models are optimized for fixed, relatively low resolutions. 
Directly applying these models to higher-resolution image editing tasks often results in quality degradation or failure, either due to resolution mismatch or prohibitive memory consumption.}

Currently, there are primarily two approaches to tackle high-resolution editing with diffusion models:
1) training a model from scratch~\cite{lugmayr2022repaint} or fine-tuning~\cite{cheng2024resadapter,zheng2024any} an existing one to higher resolutions; and 2) combining an image editing model with {an image enhancement module (e.g., post-editing via super-resolution)}. The first approach demands {substantial computational resource and training time}, even just for fine-tuning. 
For instance, training Stable Diffusion 1.5~\cite{rombach2022high} for {$512\times512$} resolution requires 20 days on 256 A100 GPUs.
{As model and dataset sizes continue to grow, this route becomes increasingly resource-intensive and time-consuming.}
The second approach, {by contrast, often suffers from poor integration between the editing and enhancement stages} (see Fig.~\ref{fig:fig_show_demo}(b)). {Moreover, the final quality heavily depends on the strength}
of the enhancement model.
{Since the editing is performed on a downsampled version of the image, the final output frequently lacks high-frequency details and visual fidelity.}

Because training models at higher resolutions is computationally expensive, and collecting high-resolution images is costly, directly extending pre-trained models to higher resolutions without {additional} training offers an appealing alternative. 
However, there  {has been limited research on adapting existing generative or editing models to ultra-high-resolution inputs}.
Existing methods \cite{bar2023multidiffusion} primarily {target high-resolution content generation from scratch,} such as the recent state-of-the-art (SOTA) method DemoFusion \cite{du2023demofusion}. 
{As shown in Fig.~\ref{fig:fig_show_demo}(c), while DemoFusion excels at generating semantically coherent high-resolution objects, it falls short in real-image editing tasks—where preserving unedited regions and maintaining global consistency are essential for practical, incremental editing workflows.}

To address these issues, we first recognize that information loss in unedited regions occurs when {latent feature maps are disrupted by random noise or overly dominant conditioning inputs—such as text prompts—that override the diffusion dynamics and inadvertently alter the appearance of unedited areas}.
The second issue is boundary artifacts between edited and unedited regions. 
Existing methods~\cite{bar2023multidiffusion,du2023demofusion} 
typically regenerate the entire latent feature map without {differentiating between edited and preserved areas, which can result in progressive detail loss across multiple diffusion steps.}
The third issue {concerns} the sampling process. High-resolution images 
{challenge both local-patch-based approaches~\cite{bar2023multidiffusion} and global dilated sampling methods~\cite{du2023demofusion}}, which often {fail to capture intermediate-scale features or sufficient receptive fields. This limitation forces a trade-off: models either overemphasize local detail or prioritize global structure, often leading to denoising artifacts and reduced visual coherence.}

We thus present UltraDiffEdit, a novel tuning-free framework for high-resolution image editing using LDMs. 
{To address the first issue, we propose a multi-scale progressive editing framework that performs iterative refinement from low to high resolution through an ``encode–diffuse–denoise–decode–blend'' loop. This design effectively fuses latent feature maps from both the original image and the generated content, enabling coarse-to-fine refinement of high-resolution outputs.}
We introduce multi-patch encoding to effectively utilize the information from edited images to maintain both edited and unedited features within the latent space. 
{To address the second issue, we design a global-local consistency denoising technique that reduces editing artifacts, preserves background content, and ensures smooth transition at the boundaries between edited and unedited regions.
Finally, to address the third issue, we propose a hybrid patch-based sampling strategy that captures local, intermediate, and global features in the latent space, thereby enhancing semantic coherence and fine detail during the sampling process}, as shown in Fig.~\ref{fig:fig_show_demo}(d).

Our approach leverages existing LDM-based generative models like Stable Diffusion series (1.5 and 2.0~\cite{rombach2022high}, and SDXL~\cite{podell2023sdxl}), ControlNet~\cite{zhang2023adding}, and IP-Adapater~\cite{ye2023ip}, {enabling seamless ultra-high-resolution real-image editing without any additional training}.
We created three benchmark datasets based on existing~\cite{DIV2K,Xie_2022_CVPR} or synthesized images, including DIV2KEdit, Syn2KEdit, and UHRSDEdit. 
{Extensive experiments and comparisons with state-of-the-art methods demonstrate the effectiveness of our approach in delivering high-quality edits at ultra-high resolutions}.

{
In summary, our paper makes four key contributions:
\begin{itemize}
\item We introduce a tuning-free framework that extends off-the-shelf LDMs to high-quality, ultra-high-resolution image editing, refining visual content in a coarse-to-fine manner {while operating under limited memory constraints.}
\item We propose {multi-patch encoding} and global-local consistency denoising to {address}  {semantic consistency and} boundary artifacts, consistently blending latent features from the generated content and the edited image.
\item We introduce a {hybrid patch-based sampling strategy} to maintain global semantic coherence and enhance detail during {denoising, incorporating a novel patch-based upsample guidance mechanism for high-resolution refinement.}
\item We built three benchmark datasets (DIV2KEdit, Syn2KEdit, and UHRSDEdit) and conducted various experiments on existing pre-trained LDMs (e.g., SDXL, ControlNet), {demonstrating ultra-high-resolution editing performance across multiple conditional modalities.}
\end{itemize}
}

\begin{figure*}[t]
    \includegraphics[width=1\textwidth]{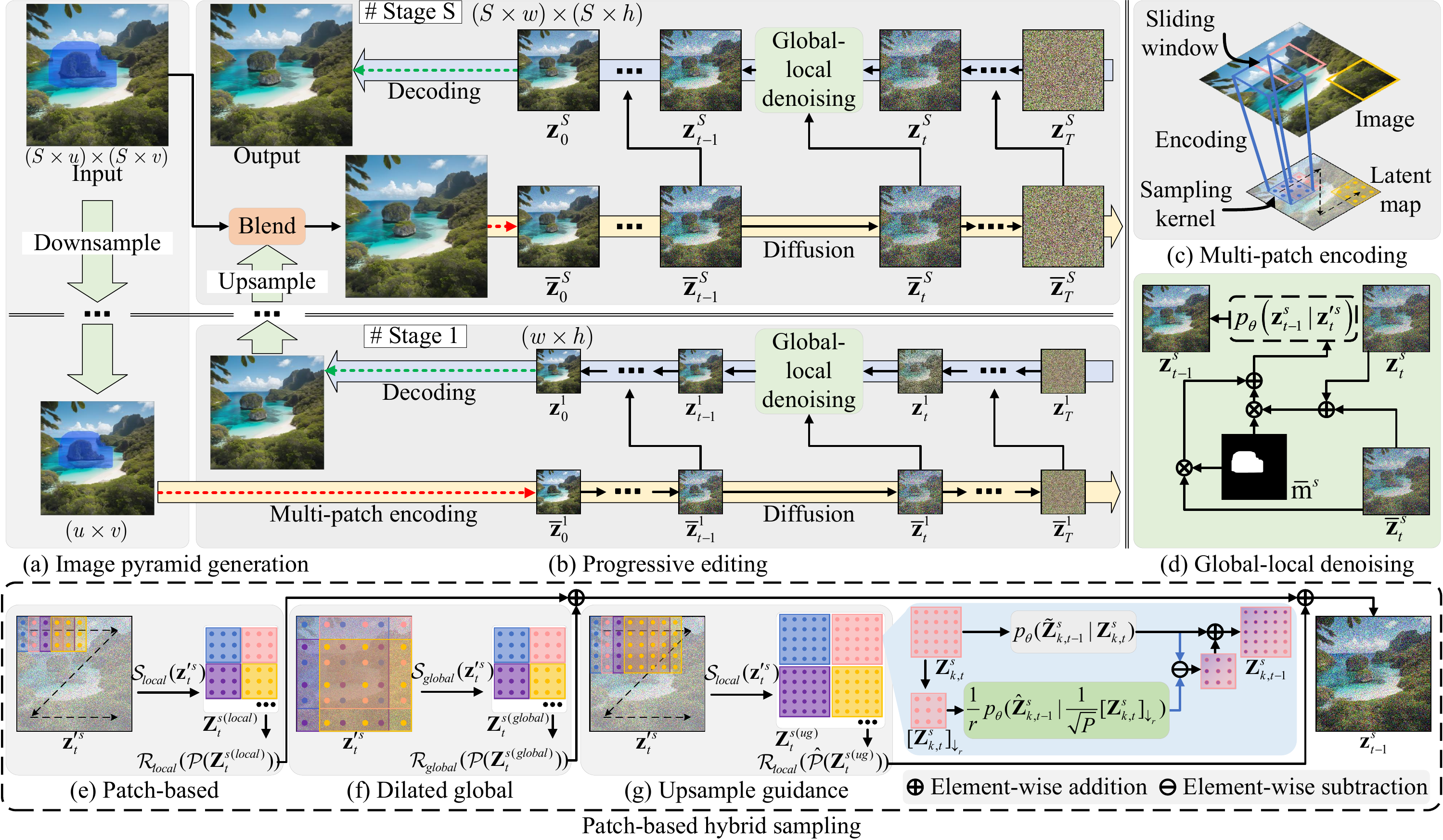}
     \caption{
    {The overall pipeline of our UltraDiffEdit. (a) We downsample an input image to create $|\mathcal{G}|$ multi-scale input images and masks (shown in blue). (b) Starting from the smallest scale, we apply progressive editing to edit the image in a coarse-to-fine manner.
     (c) Our multi-patch encoding encodes fixed-size patches for high-resolution images, merging them into a large-scale latent feature map. 
     (d) After the diffusion process, our global-local consistency denoising fuses edited and unedited latent features to ensure consistent integration across regions.
     During denoising, our patch-based hybrid sampling integrates patch-based local sampling (e), dilated global sampling (f), upsample guidance (UG) sampling (g), capturing global semantics and local details for high-quality editing. The input mask, image, and output image are blended for each scale to preserve unedited content.
    } }
	\label{fig:fig_framework}
\end{figure*}

\section{Related Work}
\textbf{High-resolution image editing.}
Image editing aims to modify the appearance, structure, or content in images, including adding or removing objects~\cite{lu2023grig}, replacing backgrounds~\cite{xie2024anywhere}, altering textures~\cite{huang2024diffusion} and expressions~\cite{FACEMUG2024}.
To achieve high-quality editing results on high-resolution images, some methods train a new model from scratch~\cite{yi2020contextual, zeng2022aggregated, CoordFill_aaai} or fine-tuning~\cite{cheng2024resadapter,zheng2024any,zhang2025diffusion} an existing one on high-resolution datasets. 
However, increasing the resolution leads to higher training costs, making existing models less practical for higher resolutions.
For instance, LDM-based models like Stable Diffusion 1.5 and Stable Diffusion XL (SDXL)~\cite{podell2023sdxl} are restricted to generating images at resolutions of $512^2$ and $1024^2$ pixels, respectively. 
Using lightweight networks~\cite{lu2023grig} can alleviate this problem but will limit the model capacity.
Another path is to combine an image editing model~\cite{manukyan2023hd,ackermann2022high,avrahami2022blended} with image enhancement techniques such as applying super-resolution post-editing~\cite{wang2021real,rombach2022high}.
However, this approach has limitations if the models are not well-integrated, and the effectiveness of enhancement is heavily influenced by the quality of the enhancement model.
To overcome these limitations, we extend pre-trained LDMs without additional training or fine-tuning.

\textbf{Image synthesis/editing using diffusion models.}
Diffusion models (DMs)~\cite{sohl2015deep, song2020denoising, ho2020denoising} have demonstrated impressive quality and diversity in image synthesis, using parameterized Markov chains to optimize the lower variational bound on the likelihood function. 
To cut computing costs, latent diffusion models (LDMs)~\cite{rombach2022high} have been introduced to perform diffusion and denoising in latent space instead of pixel space, enhancing their generalization capability and prompting further research topics, such as image inpainting~\cite{lugmayr2022repaint, corneanu2024latentpaint,grechka2024gradpaint,yang2023uni,xie2023smartbrush}, controllable generation~\cite{zhang2023adding, ye2023ip, huang2023t2i}, 
{model distillation~\cite{A_SDM_TNNLS}, high-resolution adaptation~\cite{yu2025ultraresolution}}, editable generation methods~\cite{huang2024diffusion,Kawar_2023_CVPR, zhang2023sine, wu2023latent,alharbi2024laspa,cao2023masactrl, han2024proxedit}, {and image editing benchmarks~\cite{yu2025anyedit,zhang2025diffusion}}.
However, these methods are typically trained for specific resolutions and can only produce impressive results when editing images at fixed resolutions. 
When working with higher-resolution images, they often struggle to achieve good editing performance or fail due to excessive memory demands.
This paper explores ways to adapt pre-trained generative models for high-resolution real-image editing, aiming to overcome the limitations imposed by fixed resolutions.

\textbf{Tuning-free high-resolution generation.}
Instead of developing separate models for generating images of varying resolutions, 
{high-resolution tuning-free methods~\cite{he2023scalecrafter, jin2024training,du2024max,kim2024beyondscene,wang2024generative,yang2025rectifiedhr}} try to synthesize higher-resolution images using models trained on lower-resolution images.
The intuitive idea is to generate low-resolution images and stitch~\cite{du2023demofusion,tragakis2024one} or {guide~\cite{lee2023syncdiffusion,haji2024elasticdiffusion,hwang2024upsample,guo2024make,feng2025dit4edit}} them into high-resolution images. 
MultiDiffusion~\cite{bar2023multidiffusion} collects overlapped crops from a large-size latent feature map and applies denoising processes on these crops with a shared condition input.
Several methods further add dense local descriptions~\cite{shi2024resmaster,liu2024hiprompt,lin2025accdiffusion,yang2024mastering} for overlapped crops to improve detail accuracy.
Some methods explore adjusting convolution kernels at specific layers using {re-dilation~\cite{he2023scalecrafter,wu2025megafusion}} and low-pass techniques~\cite{huang2024fouriscale,kim2024diffusehigh,si2024freeu} or manipulating the Transformer tokens~\cite{smith2024todo,bolya2023token,zhang2024HiDiffusion}.
DemoFusion~\cite{du2023demofusion} and AccDiffusion~\cite{lin2025accdiffusion} have demonstrated better generation quality, by utilizing local patch and global dilated samplings to generate higher-resolution images progressively.
However, they have limitations in high-resolution real-image editing due to random noise influencing the denoising process. This causes progressive alteration of latent feature maps during diffusion, resulting in inconsistency around editing boundaries.
In contrast, our UltraDiffEdit successfully edits high-resolution images up to 8K while ensuring overall coherence with faithful details.

\section{Methodology}

\revise{UltraDiffEdit is a tuning-free framework that extends pre-trained Latent Diffusion Models (LDMs) for ultra-high-resolution image editing. Its main idea is to progressively refine the editing result in a coarse-to-fine manner. By utilizing lower-resolution outputs and unedited regions as robust structural references for the next scale, this strategy allows the model to process ultra-high-resolution inputs within standard GPU memory constraints. 
We first introduce a multi-scale progressive editing framework, which incorporates \textit{multi-patch encoding} to preserve high-frequency details, \textit{global-local consistency denoising} to prevent structural inconsistency, and \textit{patch-based hybrid sampling} to achieve local fidelity and global coherence simultaneously.
Before detailing our modules, we briefly review the standard formulation of LDMs.}

\noindent\textbf{Latent diffusion model.} 
Given an image $\mathbf{x} \in \mathbb{R}^{3 \times u \times v}$, a pre-trained encoder $E(\cdot)$ first encodes $\mathbf{x} $ to the latent space, as $\mathbf{z}_0 = E(\mathbf{x}) \in \mathbb{R}^{c \times h \times w}$.
Then, the diffusion and denoising processes are applied to sample new data for generation. 
The \textbf{diffusion process} incrementally adds Gaussian noise to the latent feature map $\mathbf{z}_0$ and corrupts it into an approximately pure Gaussian noise  $\mathbf{z}_T$ using a variance schedule $\beta_1,\ldots, \beta_T$: $q(\mathbf{z}_{1:T}|\mathbf{z}_0) = \prod_{t=1}^{T} q(\mathbf{z}_t|\mathbf{z}_{t-1})$, $q(\mathbf{z}_t|\mathbf{z}_{t-1}) = \mathcal{N}(\mathbf{z}_t; \sqrt{1 - \beta_{t}} \mathbf{z}_{t-1}, \beta_t \mathbf{I})$.

The \textbf{denoising process} aims to recover the cleaner version $\mathbf{z}_{t-1}$ from $\mathbf{z}_{t}$ by estimating the noise, which can be expressed as $p_{\theta}(\mathbf{z}_{0:T}) = p(\mathbf{z}_T) \prod_{t=1}^{T} p_{\theta}(\mathbf{z}_{t-1} | \mathbf{z}_t)$, $p_{\theta}(\mathbf{z}_{t-1}|\mathbf{z}_t) = \mathcal{N}(\mathbf{z}_{t-1}; \mu_{\theta}(\mathbf{z}_t, t), \Sigma_{\theta}(\mathbf{z}_t, t))$,
where $\mu_{\theta}$ and $\Sigma_{\theta}$ are learned Gaussian transition, and ${\theta}$ denotes the parameters of the denoising model.

\subsection{Multi-scale progressive editing}
Fig.~\ref{fig:fig_framework} shows the UltraDiffEdit pipeline, which divides the process into $|\mathcal{G}|$ stages, each following an ``encode-diffuse-denoise-decode-blend" operation. $\mathcal{G}$ is the phase set that controls stages for editing.
A \textit{multi-scale progressive editing} is proposed to edit a given image in a coarse-to-fine manner. Our \textit{multi-patch encoding} projects a high-resolution image into LDM latent space, preserving edited and unedited latent visual details.
To eliminate artifacts at editing edges, \textit{global-local consistency denoising} consistently merges edited and unedited latent features, transitioning smoothly around edited boundaries.
Meanwhile, our \textit{patch-based hybrid sampling} maintains global semantic coherence and enhances details during denoising.

\textbf{Image pyramid generation.}
{As shown in Fig.~\ref{fig:fig_framework}(a)}, suppose we have a pre-trained image encoder $E(\cdot)$, operating on the image space $\mathbb{R}^{3 \times u \times v}$; 
and a latent diffusion model with parameters $\theta$, operating on the latent space $\mathbb{R}^{c \times h \times w}$.
Given a high-resolution image $\mathbf{X}\in \mathbb{R}^{3 \times U \times V}$ and a binary mask $\mathbf{M} \in \mathbb{R}^{1 \times U \times V}$ (with 1 for editing and 0 for unedited parts), we first downsample the input image and mask to get a sequence of downsampled images $[\mathbf{x}^1,\ldots,\mathbf{x}^s,\ldots,\mathbf{x}^S]$ and binary masks $[\mathbf{m}^1,\ldots,\mathbf{m}^s,\ldots,\mathbf{m}^S]$, where $\mathbf{x}^S = \mathbf{X}$, $\mathbf{m}^S = \mathbf{M}$, $U = S \times u$, $V = S \times v$, $s \in [1, \ldots, S]$, and $S$ is the scaling factor for the side length.

\textbf{Progressive editing.} {As shown in Fig.~\ref{fig:fig_framework}(b)}, LDM-based methods first project an input image into latent space. Using our multi-patch encoding, we can obtain a sequence of projected latent feature maps, $[\bar{\mathbf{z}}_0^{1},\ldots,\bar{\mathbf{z}}_0^{s},\ldots,\bar{\mathbf{z}}_0^{S}]$, where $\bar{\mathbf{z}}_0^{s} \in \mathbb{R}^{c \times ( s\times h) \times (s \times w)}$.
Given diffusion and denoising processes as $q(\mathbf{z}_T|\mathbf{z}_0) = \prod_{t=1}^{T} q(\mathbf{z}_t|\mathbf{z}_{t-1})$ and $p_{\theta}(\mathbf{z}_0|\mathbf{z}_T) = \prod_{t=T}^{1} p_{\theta}(\mathbf{z}_{t-1}|\mathbf{z}_t)$ from the pre-trained diffusion model with the pre-defined input size, 
we extend them to formulate the proposed multi-scale progressive editing process as:
\begin{equation}\label{equ:multi_progress_edit}
    \begin{aligned}
    p_{\theta}(\mathbf{z}_0^{S}|\bar{\mathbf{z}}_0^{1}) =\prod_{s\in \mathcal{G}} q\left(\bar{\mathbf{z}}_T^{s} | \bar{\mathbf{z}}_0^{s}\right) p_{\theta}\left(\mathbf{z}_0^{s} | \bar{\mathbf{z}}_{T}^{s} \right), 
    \end{aligned}
\end{equation}
where  $\mathcal{G}$ is the phase set that controls stages and we set $\mathcal{G}=\{1,S\}$ in this paper;
$\bar{\mathbf{z}}_0^{s} \in \mathbb{R}^{c \times (s\times h) \times (s\times w)}$ is obtained through multi-patch encoding function as:
\begin{equation}\label{equ:img_encode}
	\begin{aligned}
		\bar{\mathbf{z}}_0^{s}= \begin{cases} \mathcal{ME}(\mathbf{x}^1), & \text {if } s = 1;\\
            \mathcal{ME}([\bar{\mathbf{x}}^{\operatorname{prev}(s)}]_{{\uparrow}_{s}}), & \text {if } s \geq 2,\end{cases}
	\end{aligned}
\end{equation}
where $\bar{\mathbf{x}}^s \in \mathbb{R}^{3 \times (s\times u) \times (s\times v)}$ is the edited image at stage $s$ using pixel-level blending function, which is expressed as:
\begin{equation}\label{equ:pixel_fusion}
    \begin{aligned}
      \bar{\mathbf{x}}^s = \mathcal{D}(\mathbf{z}_0^{s}) \odot  \mathbf{m}^s  +  \mathbf{x}^s \odot  (\mathbf{1} - \mathbf{m}^s),
    \end{aligned}
\end{equation}
where $[\cdot]_{\uparrow_{s}}$ is the upsampling operation, upscaling the input to target scale $s$; $\operatorname{prev}(\cdot)$ returns the previous scale; $\mathcal{D}(\cdot)$ is the tiled decoding function~\cite{he2023scalecrafter}, generating an image $\hat{\mathbf{x}}^s \in \mathbb{R}^{3 \times (s\times u) \times (s\times v)} $ from a denoised latent feature map $\mathbf{z}_0^{s}$.

For each editing stage $s$:
(1) We encode an input image ($s=1$) or edited image ($s>1$) to get encoded feature maps $\bar{\mathbf{z}}_0^{s}$ using \textit{multi-patch encoding} to preserve unedited visual features for each stage, as Eq.~\ref{equ:img_encode}.
(2) We perform the diffusion and denoising process as $q\left(\bar{\mathbf{z}}_T^{s} | \bar{\mathbf{z}}_0^{s}\right) p_{\theta}\left(\mathbf{z}_0^{s} | \bar{\mathbf{z}}_{T}^{s} \right)$ to get the denoised latent features $\mathbf{z}_0^{s}$. Meanwhile, the \textit{global-local consistency denoising} and  \textit{patch-based hybrid sampling} are employed to enhance the semantic coherence and details in latent space.
(3) We decode the latent feature map $\mathbf{z}_0^{s}$ and blend it with unedited regions as Eq.~\ref{equ:pixel_fusion}.
At the final stage $s=S$, Poisson blending~\cite{poisson_blending} was employed to the edited image $\bar{\mathbf{x}}^S \in \mathbb{R}^{3 \times U \times V}$ to obtain a more seamless result.

\subsection{Multi-patch encoding} 
Encoding the entire image is a simple way to preserve information about both edited and unedited features in the latent space. 
However, this can result in out-of-memory errors and may not guarantee quality for very high-resolution inputs that exceed the model's trained resolution. 
Moreover, directly downsampling images will cause visual details to be lost.
Thus, we split a high-resolution image into smaller patches, encode each into the latent space, and then merge these separate embeddings to form a complete latent tensor.

{As shown in Fig.~\ref{fig:fig_framework}(c)}, given an edited image $\bar{\mathbf{x}}^s\in \mathbb{R}^{3 \times (s \times u) \times (s \times v)}$ and a binary mask $\mathbf{m}^s\in \mathbb{R}^{1 \times(s \times u) \times (s \times v)}$ at editing stage $s$, the multi-patch encoding $\mathcal{ME}(\cdot)$ can be expressed as:
\begin{equation}\label{equ:mul_patch_encode}
    \begin{aligned}
      \bar{\mathbf{z}}_0^{s} = \mathcal{R}_{local}(\mathcal{E}(S_{local}(\bar{\mathbf{x}}^s))) = \mathcal{ME}(\bar{\mathbf{x}}^s) ,
    \end{aligned}
\end{equation}
where $\mathbf{X}^\dagger= [\mathbf{X}_1,\ldots,\mathbf{X}_l,\ldots,\mathbf{X}_L] = S_{local}(\bar{\mathbf{x}}^s)$, $\mathbf{X}_l \in \mathbb{R}^{3 \times u \times v}$, and $S_{local}(\cdot)$ is a shifted crop sampling function, which performs shifted crop sampling on image space; $L = \left(\left\lfloor \frac{(s \times u)-u}{d_u} \right\rfloor+ 1\right) \times \left(\left\lfloor\frac{(s \times v)-v}{d_v} \right\rfloor + 1\right)$, $d_u$ and $d_v$ represent the stride in the vertical and horizontal directions, respectively. 
Then, we traverse all the patches in $\mathbf{X}^\dagger$ to get a sequence of latent feature maps $\mathbf{Z}^\dagger = [{\mathbf{Z}}_{1}, \ldots, {\mathbf{Z}}_{l}, \ldots, {\mathbf{Z}}_{L}] = [E(\mathbf{X}_{1}), \ldots, E(\mathbf{X}_{l}), \ldots, E(\mathbf{X}_{L})] = \mathcal{E}(\mathbf{X}^\dagger)$, where ${\mathbf{Z}}_{l} \in \mathbb{R}^{c \times h \times w}$.
To unite $\mathbf{Z}^\dagger$ to its corresponding original dimensions, we average overlapping regions and to get the latent feature map $\bar{\mathbf{z}}^s_{0} = \mathcal{R}_{local}(\mathbf{Z}^\dagger) \in \mathbb{R}^{c \times ( s\times h) \times (s \times w)}$. 
The function $\mathcal{R}_{local}(\cdot)$ averages overlapping encoded sub-latent feature maps to a large-scale latent feature map $\bar{\mathbf{z}}_0^{s}$.

\subsection{Global-local consistency denoising}
{As shown in Fig.~\ref{fig:fig_framework}(d)}, to maintain consistency between unedited and original images during denoising,
some studies focused on developing inversion techniques~\cite{hertz2023prompttoprompt,Mokady_2023_CVPR} to solve this problem. 
However, these techniques are less practical in high-resolution scenarios, due to the constraints of limited GPU memory. 

Directly diffusing $\bar{\mathbf{z}}_0^{s}$ to $\bar{\mathbf{z}}_T^{s}$ as initialization, and then denoising $\bar{\mathbf{z}}_T^{s}$ to ${\mathbf{z}}_0^{s}$ will result in most information loss of the original latent features  $\bar{\mathbf{z}}_0^{s}$. The skip residual connection was proposed in DemoFusion~\cite{du2023demofusion} as a weighted fusion between diffused and denoised latent features. However, it often distorts unedited sections due to the simple fusion of all spatial regions regardless of whether they are edited or unedited. 
Thus, we introduce a boundary-aware denoising technique that maintains the original image's background while progressively enhancing detail for edited parts.

To preserve the unedited parts and gradually introduce the generated content to fill the editing region, 
we modify $p_{\theta}\left(\mathbf{z}_{t-1}^{s} | {\mathbf{z}}_{t}^{s} \right)$ with $p_{\theta}\left(\mathbf{z}_{t-1}^{s} | {\mathbf{z}'}_{t}^{s} \right)$.
It can be expressed as:
\begin{equation}\label{equ:content_replace}
    \begin{aligned}
    \hat{\mathbf{z}}_t^s =& \gamma_1 \times \bar{\mathbf{z}}_t^s + (1 - \gamma_1) \times \mathbf{z}_t^s,\\
    {\mathbf{z}'}_t^{s} =& \hat{\mathbf{z}}_t^{s} \odot \bar{\mathbf{m}}^s + \bar{\mathbf{z}}_t^{s} \odot (\mathbf{1} - \bar{\mathbf{m}}^s),
    \end{aligned}
\end{equation}
where $\gamma_1 = ((1 + \cos(\frac{T-t}{T} \times \pi ))/2)^{{\beta}_1}$ is a scaled cosine decay factor with a scaling factor $\beta_1$; 
{${\mathbf{z}}_t^{s}$ is the output from the previous patch-based hybrid sampling step and ${\mathbf{z}'}_t^{s}$ is the input to the next patch-based hybrid sampling step;}
$\bar{\mathbf{m}}^s \in \mathbb{R}^{1 \times (s\times h) \times (s\times w)}$ is obtained by downsampling ${\mathbf{m}}^s$ to match latent map size.
{The global-local consistency denoising preserves unedited regions while smoothly introducing edited content, preventing abrupt transition. The decaying factor $\gamma_1$ progressively adjusts the fusion strength, while the mask blending ensures smooth transition along editing boundaries.}
After the $T$ step boundary-aware diffusion process, we can get the denoised feature map ${\mathbf{z}}_0^{s}$, which can be used to further generate the edited image $\bar{\mathbf{x}}^{s}$.

\subsection{Patch-based hybrid sampling} 
For denoising on high-resolution feature maps, existing local-patch~\cite{bar2023multidiffusion} and global dilated sampling~\cite{du2023demofusion} methods effectively partition and denoise patches for integration.
However, sorely combining local and global patches with limited patch size can overlook intermediate-scale features, leading to a compromise between local details and global structures, which results in denoising artifacts.

We propose a patch-based hybrid sampling that integrates local, intermediate, and global features in latent space. 
This approach enriches the receptive field during sampling, improving both global structure and local details.
The hybrid sampling can be expressed as:
\begin{equation}\label{equ:sampling}
    \begin{aligned}
    {\mathbf{z}}^s_{t-1} = (\frac{2-\gamma_2}{4}) \mathbf{z}_{t-1}^{s(local)} +  (\frac{2-\gamma_2}{4}) \mathbf{z}_{t-1}^{s(ug)} + \frac{\gamma_2}{2} \mathbf{z}_{t-1}^{s(global)},
    \end{aligned}
\end{equation}
where ${\mathbf{z}}^s_{t-1} \in \mathbb{R}^{c \times (s\times h) \times (s\times w)}$ is the denoised latent feature map; $\mathbf{z}_{t-1}^{s(local)}$,  $\mathbf{z}_{t-1}^{s(ug)}$, and $\mathbf{z}_{t-1}^{s(global)}$ are the denoised local,  upsample-guidance, and the global latent feature maps, respectively; 
$\gamma_2 = ((1 + \cos(\frac{T-t}{T} \times \pi ))/2)^{{\beta}_2}$ is a scaled cosine decay factor with a scaling factor $\beta_2$.
$\gamma_2$ prioritizes global features at earlier steps and shifts focus to detail refinement in later stages.

\textbf{Patch-based local sampling.} 
As shown in Fig.~\ref{fig:fig_framework}{(e)}, the patch-based local sampling~\cite{bar2023multidiffusion}, defined as $\mathbf{z}_{t-1}^{s(local)} = \mathcal{R}_{local}(\mathcal{P}(\mathcal{S}_{local}({\mathbf{z}'}_t^{s})))$, where $\mathcal{P}(\cdot)$ denotes the denoising process, which performs denoising for each patch.
This approach splits the large-scale latent feature map into smaller, partially overlapping patches, and denoising each separately. 
The patches are then merged by averaging the overlapping regions to reconstruct the latent tensor.

In detail, we first obtain a sequence of local latent feature maps, which is $\mathbf{Z}^{s(local)}_t = [\mathbf{Z}^s_{1,t}, \ldots, \mathbf{Z}^s_{n,t}, \ldots, \mathbf{Z}^s_{N,t}] = \mathcal{S}_{local}({\mathbf{z}'}_t^{s})$, $\mathbf{Z}^s_{n,t} \in \mathbb{R}^{c \times h \times w}$, where $N = \left(\left\lfloor \frac{(s\times h)-h}{d_h} \right\rfloor+ 1\right) \times \left(\left\lfloor\frac{(s\times w)-w}{d_w} \right\rfloor + 1\right)$, $d_h$ and $d_w$ represent the stride in the vertical and horizontal directions, respectively.
Subsequently, each local latent feature map undergoes a denoising step on the trained resolution using $p_{\theta}(\mathbf{Z}^s_{n,t-1}|\mathbf{Z}^s_{n,t})$.
This process is defined as $\mathbf{Z}^{s(local)}_{t-1}  = [\ldots, p_{\theta}(\mathbf{Z}^s_{n,t-1}|\mathbf{Z}^s_{n,t}), \ldots] = \mathcal{P}(\mathbf{Z}^{s(local)}_t)$, $n \in [1,N]$.
The local representations $\mathbf{Z}^{s(local)}_{t-1}$ are 
reconstructed back to the original dimensions, averaging any overlapping areas, resulting in $\mathbf{z}_{t-1}^{s(local)} = \mathcal{R}_{local}(\mathbf{Z}^{s(local)}_{t-1}) \in  \mathbb{R}^{c \times (s \times h) \times (s \times w)}$.
However, this approach alone can result in a loss of global consistency.
We thus utilize dilated global sampling~\cite{du2023demofusion} for capturing global information.

\textbf{Dilated global sampling.}
As shown in Fig.~\ref{fig:fig_framework}{(f)}, the dilated global sampling~\cite{du2023demofusion} is expressed as $\mathbf{z}_{t-1}^{s(global)} =  \mathcal{R}_{global}(\mathcal{P}(\mathcal{S}_{global}({\mathbf{z}'}_t^{s})))$, where $\mathcal{S}_{global}(\cdot)$ is the dilated cropping function; $\mathcal{R}_{global}(\cdot)$ is the reconstruction function for the dilated crops; $\mathcal{P}(\cdot)$ is the denosing process for each patch. 
It uses a dynamically adjusted dilation factor to expand the receptive field and capture global structural information. 
Specifically, the dilation factor is scaled based on the size factor $s$ of the latent feature map, yielding $M = s^2$ patches.

We first apply shifted dilated sampling to obtain a sequence of global latent representations, which is $\mathbf{Z}^{s(global)}_t = [\mathbf{Z}^s_{1,t}, \ldots, \mathbf{Z}^s_{m,t}, \ldots, \mathbf{Z}^s_{M,t}] = \mathcal{S}_{global}({\mathbf{z}'}_t^{s})$, $\mathbf{Z}^s_{m,t} \in \mathbb{R}^{c \times h \times w}$. 
The dilation factor is set to be $s$, and $\mathbf{Z}^{s(global)}_{t-1}  = [\ldots, p_{\theta}(\mathbf{Z}^s_{m,t-1}|\mathbf{Z}^s_{m,t}), \ldots] = \mathcal{P}(\mathbf{Z}^{s(global)}_t)$, $m \in [1,M]$.
We thus have $\mathbf{z}_{t-1}^{s(global)} = \mathcal{R}_{global}(\mathbf{Z}^{s(global)}_{t-1}) \in  \mathbb{R}^{c \times (s \times h) \times (s \times w)}$. 
However, as the size of the input latent map increases, the corresponding dilation rate also grows, causing the sampled latent map to become sparser and potentially lose intermediate-scale features. 
Thus, we introduce our patch-based upsample guidance sampling, which enriches the receptive field and enhances global-local consistency.

\textbf{Patch-based upsample guidance sampling.}
To scale up pre-trained diffusion models for high-resolution images and enhance image quality, upsample guidance~\cite{hwang2024upsample} has shown promise.
However, upsample guidance only solves the scaling up of the pre-trained size to the higher size input, but direct processing on high resolutions still leads to out-of-memory issues due to the limited GPU memory capacity.
Thus, we propose a patch-based upsample guidance sampling to alleviate this problem while improving the semantic coherence and detail in denoised latent features.

As shown in Fig.~\ref{fig:fig_framework}{(g)}, we decompose the upsample guidance sampling into a sequence of patch-based inferences and aggregate them to get a large-size denoised tensor, which is defined as $\mathbf{z}_{t-1}^{s(ug)} = \mathcal{R}_{local}(\hat{\mathcal{P}}(\mathcal{S}_{local}({\mathbf{z}'}_t^{s})))$. 
To enlarge the receptive field, a sequence of enlarged latent feature maps are calculated as $\mathbf{Z}^{s(ug)}_t = [\mathbf{Z}^s_{1,t}, \ldots, \mathbf{Z}^s_{k,t}, \ldots, \mathbf{Z}^s_{K,t}] = {\mathcal{S}}_{local}({\mathbf{z}'}_t^{s})$, $\mathbf{Z}^s_{k,t} \in \mathbb{R}^{c \times (r\times h) \times (r \times w)}$, where $K = \left(\left\lfloor \frac{(s\times h)- (r \times h)}{r \times d_h} \right\rfloor+ 1\right) \times \left(\left\lfloor\frac{(s\times w)- (r \times w)}{r \times d_w} \right\rfloor + 1\right)$, using the kernel size $(r \times h)\times (r \times w)$, and strides $r \times d_h$ and $r \times d_w$. 
We then apply the upsample guidance denoising, $\hat{p}_{\theta}(\mathbf{Z}^s_{k,t-1}|\mathbf{Z}^s_{k,t})$:
\begin{equation}
\begin{aligned}
     \epsilon^s_{k,t-1} + w_t \left[ \frac{1}{r}  {p}_{\theta}({\hat{\mathbf{Z}}^s_{k,t-1}}|\frac{1}{\sqrt{P}} [\mathbf{Z}^s_{k,t}]_{\downarrow_{r}}) - [\epsilon^s_{k,t-1}]_{\downarrow_{r}} \right]_{\uparrow_{r}},
\end{aligned}
\end{equation}
where $\epsilon^s_{k,t-1} = {p}_{\theta}(\widetilde{\mathbf{Z}}^s_{k,t-1}|\mathbf{Z}^s_{k,t})$;
{$\hat{\mathbf{Z}}^s_{k,t-1}$ is obtained by applying denoising to the downsampled noise map $[\mathbf{Z}^s_{k,t}]_{\downarrow_{r}}$;}
$[\cdot]_{\uparrow_{r}}$ and $[\cdot]_{\downarrow_{r}}$ are the upsampling and downsampling operations with the scale factor $r$, respectively; $ P = \alpha_t + \frac{1}{r^2}(1 - \alpha_t)$, which is a function of time determined by the noise schedule $\alpha_t$, and $\alpha_t=1-\beta_t$;
$w_t$ functions as the guiding scale, which depends on each time step.
Then, we can get a sequence of upsample-guided denoised features.
The patch-based upsample guidance sampling is defined as $\mathbf{Z}^{s(ug)}_{t-1}  = [\ldots, \hat{p}_{\theta}(\mathbf{Z}^s_{k,t-1}|\mathbf{Z}^s_{k,t}), \ldots] = \hat{\mathcal{P}}(\mathbf{Z}^{s(ug)}_t)$, $k \in [1,K]$.
We thus have $\mathbf{z}_{t-1}^{s(ug)} = \mathcal{R}_{local}(\mathbf{Z}^{s(ug)}_{t-1}) \in  \mathbb{R}^{c \times (s \times h) \times (s \times w)}$.

\revise{Our UltraDiffEdit introduces three novel designs explicitly tailored for ultra-high-resolution real-image editing: (1) \textit{multi-patch encoding} that propagates decoded pixel-level references to preserve original high-frequency details at each scale; (2) \textit{global-local consistency denoising} that fuses features at every diffusion step to prevent structural inconsistency and eliminate the boundary artifacts; and (3) our \textit{patch-based hybrid sampling} that dynamically integrates multiple scales to guarantee both fine-grained local details and seamless global coherence. Together, these designs ensure seamless and structurally coherent results, directly overcoming the fundamental bottlenecks of existing high-resolution editing approaches.}

\section{Experiments}
This section presents qualitative and quantitative results, ablation studies, and applications. 
More implementation details and experimental results are available in the supplementary document.

\subsection{Implementation details}
 
{The pseudo-code of our UltraDiffEdit is provided in Algorithm~\ref{alg:ultra_diff_inference}. Given an input image $\mathbf{X}$ and a binary mask $\mathbf{M}$, an image pyramid 
$\{\mathbf{x}^s, \mathbf{m}^s\}_{s \in \mathcal{G}}$  is first constructed across multiple scales by the downsampling operation ($[\cdot]_{\downarrow_s}$). 
For each scale $s$, the latent feature $\bar{\mathbf{z}}_0^s$ is obtained via the multi-patch encoding $\mathcal{ME}(\cdot)$, initialized either from the smallest scale ($s=1$) or the upsampled result from the previous scale $[\bar{\mathbf{x}}^{\operatorname{prev}(s)}]_{\uparrow_s}$. 
During the diffusion process ($t = T \rightarrow 1$), local and global consistencies are enforced through global-local consistency denoising, 
followed by mask-guided fusion. 
Patch-based hybrid sampling is performed in three complementary spaces, which are patch-based local sampling, dilated global sampling, and patch-based upsample guidance sampling. 
The corresponding latent maps $(\mathbf{z}_{t-1}^{s(local)}$, $\mathbf{z}_{t-1}^{s(ug)}$, and $\mathbf{z}_{t-1}^{s(global)}$) are then reconstructed and fused to obtain ${\mathbf{z}}^s_{t-1}$.
Finally, the decoder $\mathcal{D}(\cdot)$ reconstructs the edited image and pixel-level blending produces the refined output $\bar{\mathbf{x}}^s$ for each scale. 
The refined output $\bar{\mathbf{x}}^S$ from the final scale serves as the result.}

\begin{algorithm}[!t]
\caption{{Multi-scale progressive editing of UltraDiffEdit}}
\label{alg:ultra_diff_inference}
    \begin{algorithmic}[1]
    
    \For{$s \in \mathcal{G}$} \hfill \textcolor{gray}{\# Image pyramid generation}
            \State $\mathbf{x}^s \leftarrow [\mathbf{X}]_{{\downarrow}_s}$ 
            \State $\mathbf{m}^s \leftarrow [\mathbf{M}]_{{\downarrow}_s}$
    \EndFor

    \For{$s \in \mathcal{G}$} \hfill \textcolor{gray}{\# Traverse editing stages}
            \State \textcolor{gray}{\# Multi-patch encoding}
            \State $\bar{\mathbf{z}}_0^{s}= \begin{cases} \mathcal{ME}(\mathbf{x}^1), & \text {if } s = 1;\\
            \mathcal{ME}([\bar{\mathbf{x}}^{\operatorname{prev}(s)}]_{{\uparrow}_{s}}), & \text {if } s \geq 2,\end{cases}$ 
           
            \State $\bar{\mathbf{m}}^{s} \leftarrow [{\mathbf{m}}^{s}]_{{\downarrow}_{\hat{s}}}$ \hfill \textcolor{gray}{\# Downsample mask}
           
            \For{$t = 1$ to $T$} \hfill \textcolor{gray}{\# Diffusion steps} 
                \State $q\left(\bar{\mathbf{z}}_{t}^{s} | \bar{\mathbf{z}}_{t-1}^{s}\right)$ 
            \EndFor
           
            \State $\mathbf{z}_T^s \sim \mathcal{N}(0, \mathbf{I})$ \hfill \textcolor{gray}{\# Random initialization}
            \For{$t = T$ to $1$}
                \State \textcolor{gray}{\# Global-local consistency denoising}
                \State $\hat{\mathbf{z}}_t^s \leftarrow \gamma_1 \times \bar{\mathbf{z}}_t^s + (1 - \gamma_1) \times \mathbf{z}_t^s$
                \State ${\mathbf{z}'}_t^{s} \leftarrow \hat{\mathbf{z}}_t^{s} \odot \bar{\mathbf{m}}^s + \bar{\mathbf{z}}_t^{s} \odot (\mathbf{1} - \bar{\mathbf{m}}^s)$ \hfill \textcolor{gray}{\# fusion}
                \State $\mathbf{Z}^{s(local)}_t \leftarrow \mathcal{S}_{local}({\mathbf{z}'}_t^{s})$ \hfill \textcolor{gray}{\# Patch-based cropping}
                \State $\mathbf{Z}^{s(global)}_t \leftarrow \mathcal{S}_{global}(\mathbf{z}'^s_t)$ \hfill \textcolor{gray}{\# Dilated cropping}
                \State \textcolor{gray}{\# Patch-based upsample guidance cropping}
                \State $\mathbf{Z}^{s(ug)}_t \leftarrow   {\mathcal{S}}_{local}(\mathbf{z}'^s_t)$
                \For{$\mathbf{Z}^s_{n,t} \in \mathbf{Z}^{s(local)}_t$} \hfill \textcolor{gray}{\# Patch-based sampling}
                    \State $p_{\theta}(\mathbf{Z}^s_{n,t-1}|\mathbf{Z}^s_{n,t})$
                \EndFor
                \For{$\mathbf{Z}^s_{m,t} \in \mathbf{Z}^{s(global)}_t$} \hfill \textcolor{gray}{\# Dilated sampling}
                    \State $p_{\theta}(\mathbf{Z}^s_{m,t-1}|\mathbf{Z}^s_{m,t})$
                \EndFor
                \State \textcolor{gray}{\# Patch-based upsample guidance sampling}
                \For{$\mathbf{Z}^s_{k,t} \in \mathbf{Z}^{s(ug)}_t$}
                    \State $\hat{p}_{\theta}(\mathbf{Z}^s_{k,t-1}|\mathbf{Z}^s_{k,t})$
                \EndFor
                \State \textcolor{gray}{\# Latent map reconstruction} 
                \State $\mathbf{z}_{t-1}^{s(local)} \leftarrow \mathcal{R}_{local}(\mathbf{Z}^{s(local)}_{t-1})$
                \State $\mathbf{z}_{t-1}^{s(global)} \leftarrow \mathcal{R}_{global}(\mathbf{Z}^{s(global)}_{t-1})$
                \State $\mathbf{z}_{t-1}^{s(ug)} \leftarrow \mathcal{R}_{local}(\mathbf{Z}^{s(ug)}_{t-1})$

                \State \textcolor{gray}{\# Patch-based hybrid sampling} 
                \State ${\mathbf{z}}^s_{t-1} \leftarrow (\frac{2-\gamma_2}{4}) (\mathbf{z}_{t-1}^{s(local)} + \mathbf{z}_{t-1}^{s(ug)}) + \frac{\gamma_2}{2} \mathbf{z}_{t-1}^{s(global)}$ 
            \EndFor
            \State \textcolor{gray}{\# Pixel-level blending} 
            \State $\bar{\mathbf{x}}^s \leftarrow \mathcal{D}(\mathbf{z}_0^{s}) \odot  \mathbf{m}^s  +  \mathbf{x}^s \odot  (\mathbf{1} - \mathbf{m}^s)$ 
    \EndFor
    \State \textbf{return} $\bar{\mathbf{x}}^S$ 
    \end{algorithmic}
\end{algorithm}

We applied our method to Stable Diffusion (SD) 1.5, 2.0~\cite{rombach2022high} (default resolution of 512$^2$), and SDXL~\cite{podell2023sdxl} (default resolution of 1024$^2$). 
Unless specified otherwise, results are visualized using SDXL, with a DDIM scheduler of 50 steps and a guidance scale of 7.5 for all denoising paths.
For multi-patch encoding, we used predefined cropping sizes: $u = 512$, $v = 512$ for SD 1.5 and 2.0, as well as $u = 1024$, $v = 1024$ for SDXL, with a stride of $d_u = u/2$ and $d_v = v/2$ in both cases. We applied random perturbations with offsets up to $u/16$ and $v/16$ to mitigate seam issues.
We set $h = 64$, $w = 64$ for SD 1.5 and 2.0, as well as $h = 128$, $w = 128$ for SDXL, matching the maximum training size of pre-trained latent diffusion models, with a stride of $d_h = h/2$ and $d_w = w/2$, and included random perturbations of $h/16$ and $w/16$ as well.
We utilized decay factors $\beta_1 = 3$, $\beta_2 = 1$; and $r = 2$, $w_t = 0.2 \times \mathbb{I}(t>500)$ for patch-based upsample guidance sampling.
For images with various aspect ratios, padding was calculated to align with the highest resolution and then cropped back to the original size after editing.

\textbf{Benchmark datasets.}
For high-resolution image editing, we developed three benchmark datasets: DIV2KEdit, Syn2KEdit, and UHRSDEdit. 
Each dataset includes corresponding images, text prompts, editing masks, sketches, depth maps, and pose keypoints for real-image editing tasks.
DIV2KEdit and UHRSDEdit are sourced from the DIV2K~\cite{DIV2K} validation set and the UHRSD~\cite{Xie_2022_CVPR} test set, respectively. 
DIV2KEdit comprises 100 images at 2K resolution, while UHRSDEdit contains 988 images with resolutions ranging from 4K to 8K.
Syn2KEdit, a synthetic dataset using ChatGPT~\cite{ChatGPT} and DemoFusion, consists of 100 images at 2K resolution featuring diverse styles, scenes, and objects.

\begin{figure*}[t]
    \centering
    \includegraphics[width=1\textwidth]{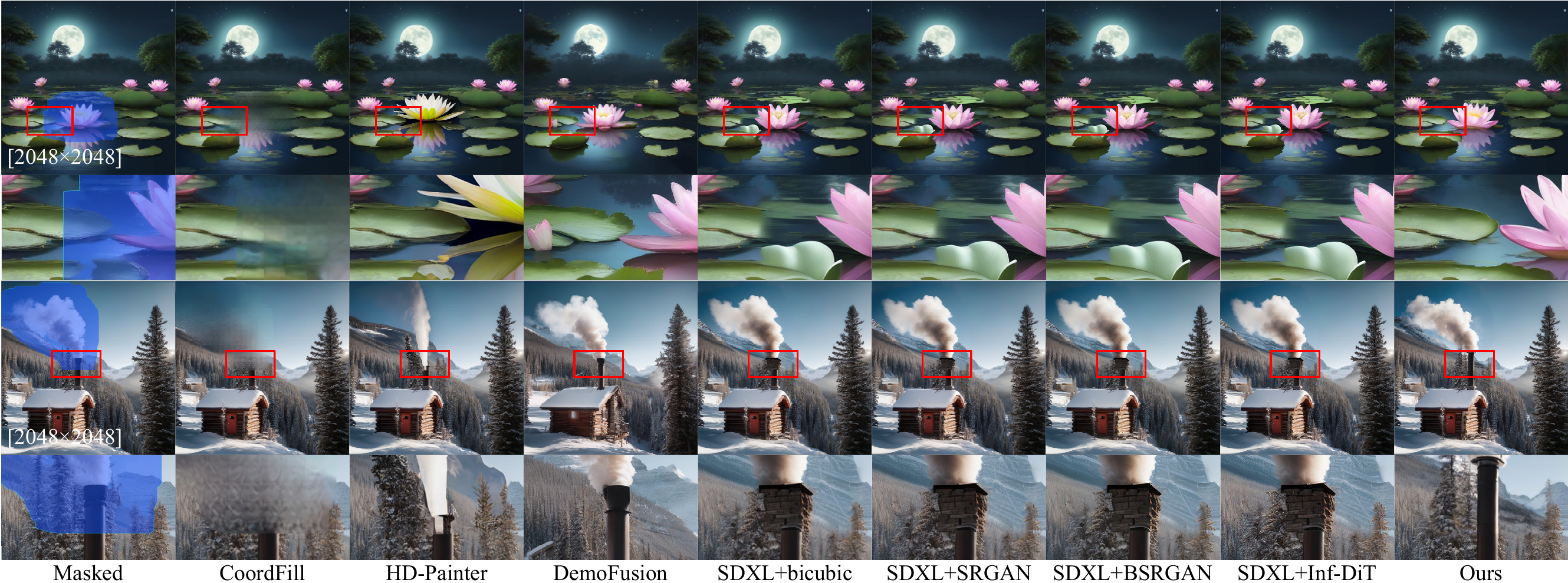}
    \caption{
    Visual results of the compared methods. Our UltraDiffEdit shows better global-local consistency with faithful details.
    } 
    \label{fig:fig_vis_compare}
\end{figure*}

\begin{table*}[htbp]
    \centering
    \caption{Quantitative results on high-resolution images. UltraDiffEdit gets superior performance on most metrics. ``N/A'': out-of-memory.}
    \label{tab:comparison}
    \resizebox{\textwidth}{!}{%
        \begin{tabular}{c|cccc|cccc|cccc}
            \hline
            \multirow{1}{*}{Dataset} &\multicolumn{4}{c|}{DIV2KEdit (2K)}  & \multicolumn{4}{c|}{Syn2KEdit (2K)} & \multicolumn{4}{c}{UHRSDEdit (4K-8K)} \\
            \hline
            \multirow{1}{*}{Metric} &PSNR$^{\uparrow}$& SSIM$^{\uparrow}$&U-IDS$_{\text{crop}}^{\uparrow}$& CLIP-S$_{\text{crop}}^{\uparrow}$ & PSNR$^{\uparrow}$& SSIM$^{\uparrow}$& U-IDS$_{\text{crop}}^{\uparrow}$ & CLIP-S$_{\text{crop}}^{\uparrow}$ & PSNR$^{\uparrow}$& SSIM$^{\uparrow}$& U-IDS$_{\text{crop}}^{\uparrow}$ & CLIP-S$_{\text{crop}}^{\uparrow}$ \\
            \hline
            CoordFill &   16.37  &  0.6934    &   0.00\% &   \textbf{23.18\%}     &    18.89  &  0.8619   & 0.00\% & \textbf{21.78\%} &  N/A  & N/A & N/A & N/A   \\
            HD-Painter &   14.65  &   0.6490   &   0.00\% &  21.22\%      &   16.86   &  0.8343   & 0.00\% & 20.66\% & N/A & N/A & N/A  & N/A\\
            DemoFusion &   15.65  &  0.4110    &  0.00\% &    22.23\%     &   19.71   &  0.7016   & 0.00\% & 21.26\% &  17.18 & 0.5996 & 6.59\% & 22.74\%  \\
            SDXL+bicubic &  17.65   &  0.5791    &  0.00\%  &   21.25\%     &   21.51 & 0.8009  &   0.00\%  &  20.67\% & 18.01 & 0.7042   &   6.58\%   &  \textbf{23.02\%}   \\
            SDXL+SRGAN &    17.53    &  0.5677   &  0.00\% & 21.20\%   & 21.37 & 0.7878 & 0.00\%  &  20.65\% &   N/A  & N/A & N/A & N/A \\
            SDXL+BSRGAN &  17.29   &  0.5649    &  0.00\%  &  21.39\%     & 21.11     &   0.7823  & 0.00\% & 20.70\% &   N/A  & N/A & N/A & N/A  \\
            SDXL+Inf-DiT  &  17.10    &  0.5398    &  0.00\% &   21.25\%      &  20.85    &  0.7522   & 0.00\% &  20.60\% & N/A  & N/A & N/A & N/A \\
            UltraDiffEdit & \textbf{18.99} & \textbf{0.7674}  & \textbf{4.58\%}& 21.40\%   & \textbf{22.39} & \textbf{0.9033} & \textbf{5.55\%} &  21.24\% &  \textbf{18.53}  & \textbf{0.8393} & \textbf{27.03\%} &  22.70\%\\ 

            \hline
            \hline
            \multirow{1}{*}{Metric} &FID$^{\downarrow}$& LPIPS$^{\downarrow}$ &FID$_{\text{crop}}^{\downarrow}$& LPIPS$_{\text{crop}}^{\downarrow}$&FID$^{\downarrow}$& LPIPS$^{\downarrow}$  & FID$_{\text{crop}}^{\downarrow}$& LPIPS$_{\text{crop}}^{\downarrow}$& FID$^{\downarrow}$& LPIPS$^{\downarrow}$ & FID$_{\text{crop}}^{\downarrow}$& LPIPS$_{\text{crop}}^{\downarrow}$ \\
            \hline
            CoordFill &   168.90  &   0.2857    &  126.64   &    0.3938    &    156.58  &   0.1852  & 78.31  &   0.2676  &   N/A  & N/A & N/A & N/A \\
            HD-Painter & 106.01    &  0.2752    &  78.47   &    0.3630    &  99.24    &   0.1737  &  55.98 &  0.2494   &  N/A  & N/A & N/A & N/A  \\
            DemoFusion &   121.90  &   0.4562   &   90.94  &   0.5103    &   73.53   &  0.2990   & 43.05 & 0.3605  & 92.57 &  0.4893 &  36.40 &  0.5637 \\
            SDXL+bicubic &  89.67   &  0.2411     & 71.06  &    0.3446      &   59.80   & 0.1221   &   41.29  &  0.2118 & \textbf{42.23} &  0.2358   &  18.76  &  0.4409  \\
            SDXL+SRGAN &    89.76    &  0.2417  & 65.23 &  0.3348  &  59.73  &  {0.1221}   &  36.77 & \textbf{0.1221} & N/A  & N/A & N/A & N/A \\
            SDXL+BSRGAN &  92.93   &   0.2532   &    69.48 & 0.3595     &   62.01   & 0.1371    &  41.96  &  0.2541    &   N/A  & N/A & N/A & N/A  \\
            SDXL+Inf-DiT  &  89.52    &  0.2456    &  65.02 &   0.3539     &  60.24    &  0.1270     & 38.90   &  0.2392 & N/A  & N/A & N/A & N/A\\
            UltraDiffEdit &  \textbf{78.59}  &  \textbf{0.2104} &   \textbf{57.65} & \textbf{0.2645}  &      \textbf{52.66} &  \textbf{0.1053} & \textbf{30.05} &{0.1512} & 42.37 & \textbf{0.2130} & \textbf{7.45} & \textbf{0.2669} \\ 
            \hline
        \end{tabular}
    }
\end{table*}

\begin{figure*}[t]
    \centering
    \includegraphics[width=1.0\textwidth]{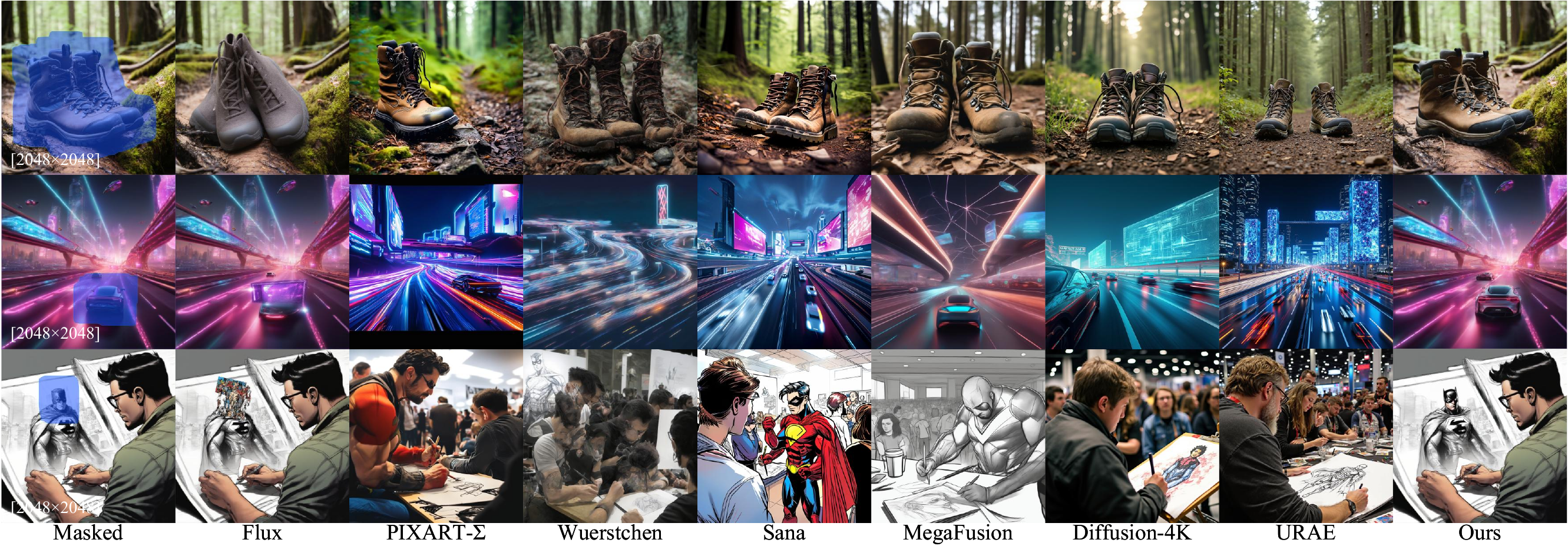}
    \caption{
    {Visual results of the compared methods. UltraDiffEdit shows better global-local consistency with high fidelity.}
    } 
    \label{fig:fig_sys2k_new_compare2_extra}
\end{figure*}

\begin{figure*}[t]
    \centering
    \includegraphics[width=1.0\textwidth]{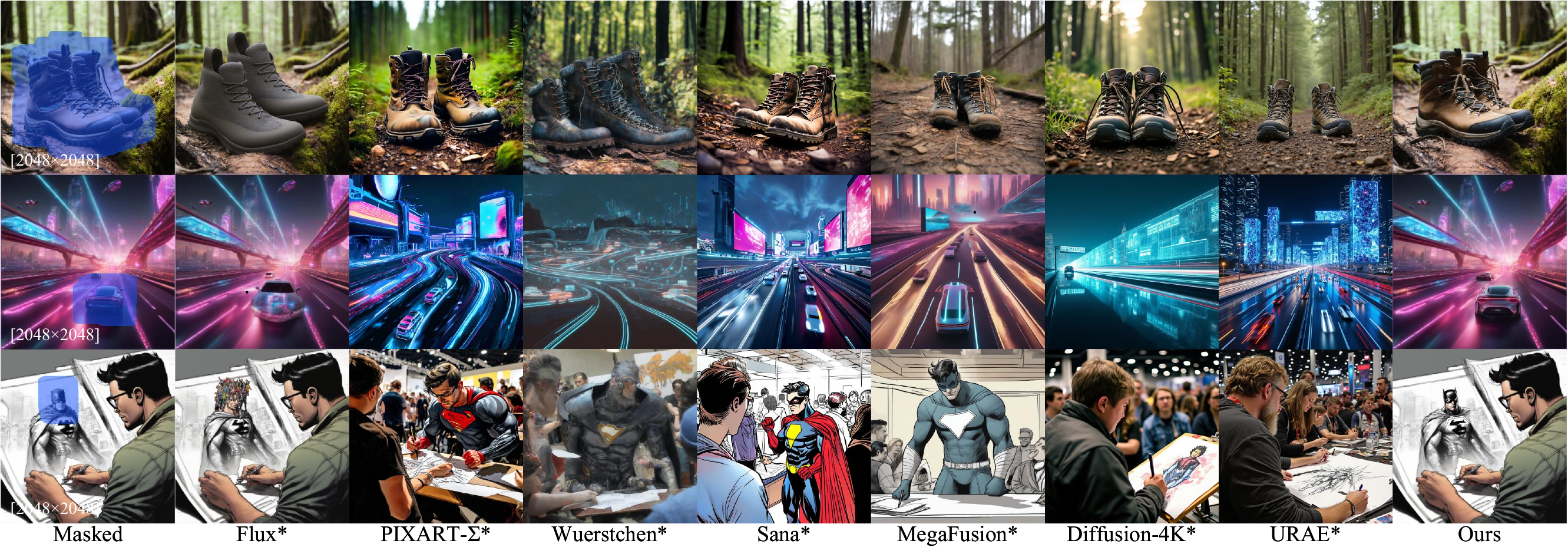}
    \caption{
    {Visual results of the compared methods. * indicates increased inference steps to match the runtime
of UltraDiffEdit.}
    } 
    \label{fig:fig_sys2k_new_compare3_extra}
\end{figure*}

\textbf{Methods for comparison and application.}
We compared our method to SOTA image inpainting and editing methods, including image inpainting method CoordFill~\cite{CoordFill_aaai}, large-scale high-resolution image editing methods (HD-Painter~\cite{manukyan2023hd}, DemoFusion~\cite{du2023demofusion}, SDXL~\cite{podell2023sdxl}+bicubic, SDXL+SRGAN~\cite{Ledig_2017_CVPR}, SDXL+BSRGAN~\cite{zhang2021designing}, and SDXL+Inf-DiT~\cite{InfDiT}). 
{
Since SDXL was not originally designed for ultra-high-resolution editing, we chose SDXL+Bicubic and SDXL+SRGAN as practical baselines, representing common strategies that upscale or refine outputs to meet high-resolution requirements.
For fair comparison, we adapted these pipelines by first generating edited images with SDXL at the 1K resolution and then applying bicubic interpolation or super-resolution models (SRGAN/BSRGAN) to upscale them to the target resolution. 
This setup provides a realistic reference for how users might employ existing models for high-resolution editing when dedicated solutions are not available.} For multimodal image editing,
we extended ControlNet~\cite{zhang2023adding} and IP-Adapter~\cite{ye2023ip} for high-resolution editing with additional conditioning inputs.

\textbf{Metrics.}
We used PSNR, SSIM, U-IDS~\cite{Zhao2021}, CLIP score (CLIP-S)~\cite{CLIP_2021}, LPIPS~\cite{Zhang2018_lpips}, and FID~\cite{Heusel2017} metrics for quantitative evaluation following established practice in recent literature.
{PSNR and SSIM measure pixel-level fidelity and structure-level fidelity against the ground truth, respectively. U-IDS evaluates perceptual fidelity, which is essential for human-related edits. CLIP score reflects semantic alignment with text prompts. LPIPS captures perceptual similarity using deep features, and FID assesses distribution-level realism of the outputs. Together, these complementary metrics provide a comprehensive evaluation of both fidelity and perceptual quality in high-resolution image editing.}
Since CLIP score, U-IDS, LPIPS, and FID require resizing images to a fixed size (224$^2$ for the CLIP score or 299$^2$ for the other three metrics), which may overlook details in high-resolution images, we also crop local patches of 1024$^2$ and resize them for these metrics, referred to as CLIP-S$_{\text{crop}}$, U-IDS$_{\text{crop}}$, LPIPS$_{\text{crop}}$, and FID$_{\text{crop}}$, respectively.

\subsection{Experimental comparisons}\label{sec:exp_comparisons}
Fig.~\ref{fig:fig_vis_compare} shows visual comparisons among different methods. 
While all approaches produce plausible edited content, inpainting techniques such as CoordFill frequently produce blurrier results and lack precise control over editing. 
Large vision models, such as HD-Painter\cite{manukyan2023hd}, deliver impressive outcomes with text prompts but may have inconsistencies between foreground and background. Upsampling methods like SDXL+BSRGAN~\cite{zhang2021designing} and SDXL+Inf-DiT~\cite{InfDiT} provide temporary high-resolution solutions but can suffer from integration issues and out-of-memory errors at 4K resolutions. 
In contrast, our UltraDiffEdit produces high-fidelity, visually coherent edits and allows for inference on ultra-high resolutions.

Fig.~\ref{fig:fig_show_demo} highlights UltraDiffEdit's performance on ultra-high-resolution images. 
Although SDXL with bicubic upsampling is an option, it yields blurry details and lacks semantic consistency. 
DemoFusion~\cite{du2023demofusion} shows strong generative abilities but fails to retain unedited content. 
Our method maintains high fidelity and semantic coherence by preserving unedited areas in latent space with multi-patch encoding and ensuring quality edits through our hybrid sampling.

Table~\ref{tab:comparison} provides quantitative comparisons on the DIV2KEdit, Syn2KEdit, and UHRSDEdit datasets. 
While all methods perform well at resolutions below 2K, the majority struggle with higher resolutions. 
Inpainting methods like CoordFill excel in pixel-level metrics (SSIM, PSNR) by preserving unedited areas but lack perceptual quality compared to larger vision models due to limited capacity.
HD-Painter~\cite{manukyan2023hd} achieves impressive results with text prompts but cannot handle resolutions above 4K. 
Upsampling methods like SDXL+BSRGAN and SDXL+Inf-DiT offer strong performance but often alter untouched areas, reducing pixel similarity. 
Most approaches fail with the 8K resolution of UHRSDEdit due to GPU memory limitations. 
In contrast, DemoFusion and UltraDiffEdit use patch-based inference to handle ultra-high resolutions.  Note that UltraDiffEdit's global-local consistency denoising produces consistent results, outperforming other methods across metrics while preserving untouched areas.

\begin{table}[t]
\setlength{\abovecaptionskip}{0.1cm}
    \centering
    \caption{Comparisons to text-to-image editing and generation. * indicates increased inference steps to match the runtime of UltraDiffEdit.}
    \label{tab:extra_compare}
    \resizebox{\linewidth}{!}{%
        \begin{tabular}{c|cccccccc}
            \hline
            \multirow{1}{*}{Metric} &PSNR$^{\uparrow}$& SSIM$^{\uparrow}$&U-IDS$_{\text{crop}}^{\uparrow}$& CLIP-S$_{\text{crop}}^{\uparrow}$ &FID$^{\downarrow}$& LPIPS$^{\downarrow}$ &FID$_{\text{crop}}^{\downarrow}$& LPIPS$_{\text{crop}}^{\downarrow}$ \\
            \hline
             Flux  &  18.67 &  0.8382 & 0.00\% &  20.68\% &   90.09  & 0.1633 &  51.28 & 0.2366  \\ 
            PIXART-$\Sigma$ &  8.82 &  0.3789 &  0.00\% & 20.72\%    & 169.74  & 0.6980 & 90.04 &0.7387\\
            Wuerstchen  &   9.99 & 0.4969  & 0.00\%   &   20.99\%  &   180.21 &    0.6636   &  94.86  & 0.6961 \\
             Sana  & 8.56 & 0.3771 & 0.00\% & 20.68\% &  169.44 & 0.6827  &  91.25 &  0.7173  \\ 
             {Megafusion} & {10.18} & {0.5116}&  {0.00\%} &  {21.06\%}  &  {164.28}   &  {0.6501} & {83.73}  &  {0.6873} \\ 
             {Diffusion-4k} & {9.24} & {0.4561} &  {0.00\%}  & {20.94\%}  &   {168.23}  & {0.6828} & {89.87} & {0.7228} \\ 
             {RectifiedHR} &  {9.85} & {0.4283} &  {0.00\%} & {20.43\%} & {162.17} & {0.6825} & {87.60} & {0.7141} \\
             {URAE} &  {9.28} & {0.4013} & {0.00\%} &  {20.78\%} & {173.56}& {0.6882} & {92.87} & {0.7152}  \\
             \hline
             {Flux*}  &  {18.42} &  {0.8268} & {0.00\%} &  {20.60\%} &   {93.59}  & {0.1666} &   {53.73} &  {0.2436}  \\ 
            {PIXART-$\Sigma$*} & {8.73} & {0.3692}  &  {0.00\%} & {20.71\%} &  {166.74} & {0.6983} & {91.00} &  {0.7340} \\
            {Wuerstchen*}  &   {10.18} & {0.4900} & {0.00\%} &   {20.73\%} & {171.23} & {0.6699}   &  {95.88} &  {0.7026} \\
            {Sana*}  & {8.54} & {0.3680} & {0.00\%} & {20.62\%} &  {166.98} & {0.6878} &  {90.28} &  {0.7236}  \\ 
             {Megafusion*} &  {10.13} & {0.5126} & {0.00\%} & {21.05\%}  &  {150.37} & {0.6537} &  {82.85} & {0.6921} \\ 
             {Diffusion-4k*} &  {9.09} & {0.4513}   & {0.00\%}  &  {20.97\%}  &  {171.67}  & {0.6894} & {91.02}  &  {0.7299}  \\ 
             {RectifiedHR*} &  {8.88} & {0.3108} & {0.00\%} &  {20.39\%} &   {196.68}  &  {0.7090}&  {108.43} &  {0.7403} \\
             {URAE*} & {9.33} & {0.4046} & {0.00\%}  & {20.83\%}  & {176.64}& {0.6835} &  {94.05} & {0.7117}\\
             \hline
            Ours &  \textbf{22.39} & \textbf{0.9033} & \textbf{5.55\%} &  \textbf{21.24\%} &\textbf{52.66} &  \textbf{0.1053} & \textbf{30.05} &\textbf{0.1512}   \\ 
            \hline
        \end{tabular}
    }
\end{table}

{
While UltraDiffEdit achieves leading performance across all datasets, its metric values vary due to differences in data characteristics. 
On Syn2KEdit, which contains synthetic images, the model achieves the highest PSNR (22.39), SSIM (0.9033), and strong LPIPS scores. This is likely because synthetic images contain less high-frequency detail, making pixel-level and perceptual reconstruction easier. 
However, its FID score (52.66) is not as competitive, possibly due to the greater style and texture diversity in synthetic data, which increases distributional variance and challenges generative alignment.
In contrast, UHRSDEdit includes higher-resolution natural images (4K–8K) and more samples (988 vs. 100), leading to increased variation in FID and U-IDS scores. The larger scale and resolution also make inference more difficult, contributing to a lower PSNR (18.53), though perceptual quality remains strong (FID 42.37, U-IDS 27.03\%). These results show that both content type and dataset scale significantly influence quantitative evaluation.
}

Fig.~\ref{fig:fig_sys2k_new_compare2_extra} and {Fig.~\ref{fig:fig_sys2k_new_compare3_extra}} present more qualitative results of the advanced high-resolution image editing (Flux~\cite{flux2024}) and generation methods (PIXART-$\Sigma$~\cite{chen2024pixart}, Wuerstchen~\cite{pernias2024wrstchen}, Sana~\cite{xie2024sana}, {Megafusion~\cite{wu2025megafusion}, Diffusion-4k~\cite{zhang2025diffusion}, RectifiedHR~\cite{yang2025rectifiedhr}, and URAE~\cite{yu2025ultraresolution}}). 
{To ensure a fair comparison in terms of runtime, we further adjust the number of inference steps for each method to approximately match the inference time of UltraDiffEdit (109.71s). 
Specifically, we set the number of steps to 55 for Flux*, 174 for PIXART-$\Sigma$*, 401 for Wuerstchen*, 499 for Sana* (499 is the maximum allowable step count for Sana*), {240 (192 in stage 1 and 48 in stage 2) for Megafusion*, 72 for Diffusion-4k*,  140 for RectifiedHR*, and 37 for URAE*}.}
It shows that each method can produce high-quality results for high-resolution inputs, but some of them (PIXART-$\Sigma$, Wuerstchen, Sana, {Megafusion, Diffusion-4k, RectifiedHR, and URAE}) cannot preserve the unedited parts for further incremental image editing. In contrast, our UltraDiffEdit achieves high-fidelity results while showing better effects on global-local consistency and faithful details.

Table~\ref{tab:extra_compare} shows more quantitative comparisons to these text-to-image editing and generation methods on the Syn2KEdit dataset (with 2K resolution).  Note that only a 2K version of PIXART-$\Sigma$ was accessible.
{As shown in the results, even when increasing the inference steps to match our runtime, their image quality remains similar (e.g., Flux vs. Flux*).}
It highlights UltraDiffEdit’s superior performance in real-image editing tasks.
Many of the referenced works primarily target text-to-image generation and do not directly address the specific challenges we tackle, such as boundary consistency and global-local coherence in real-image editing scenarios.

\begin{figure}[t]
    \centering
    \includegraphics[width=0.485\textwidth]{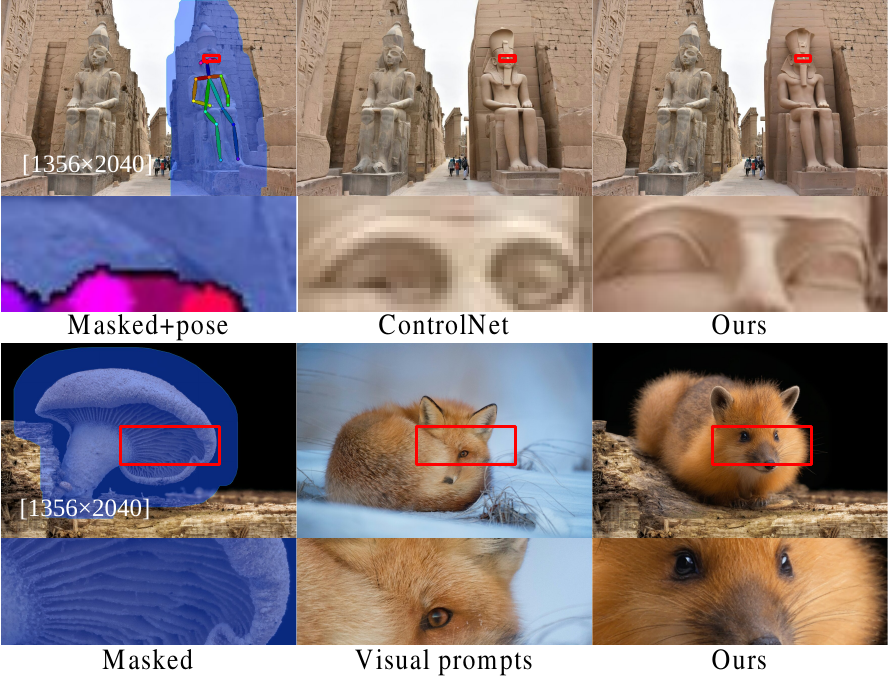}
    \caption{
    Visual results of extending existing image editing methods: ControlNet~\cite{zhang2023adding} (top) and IP-adapter~\cite{ye2023ip} (bottom).
    } 
    \label{fig:fig_vis_app_control}
\end{figure}

\begin{figure*}[!th]
    \centering
    \includegraphics[width=1\textwidth]{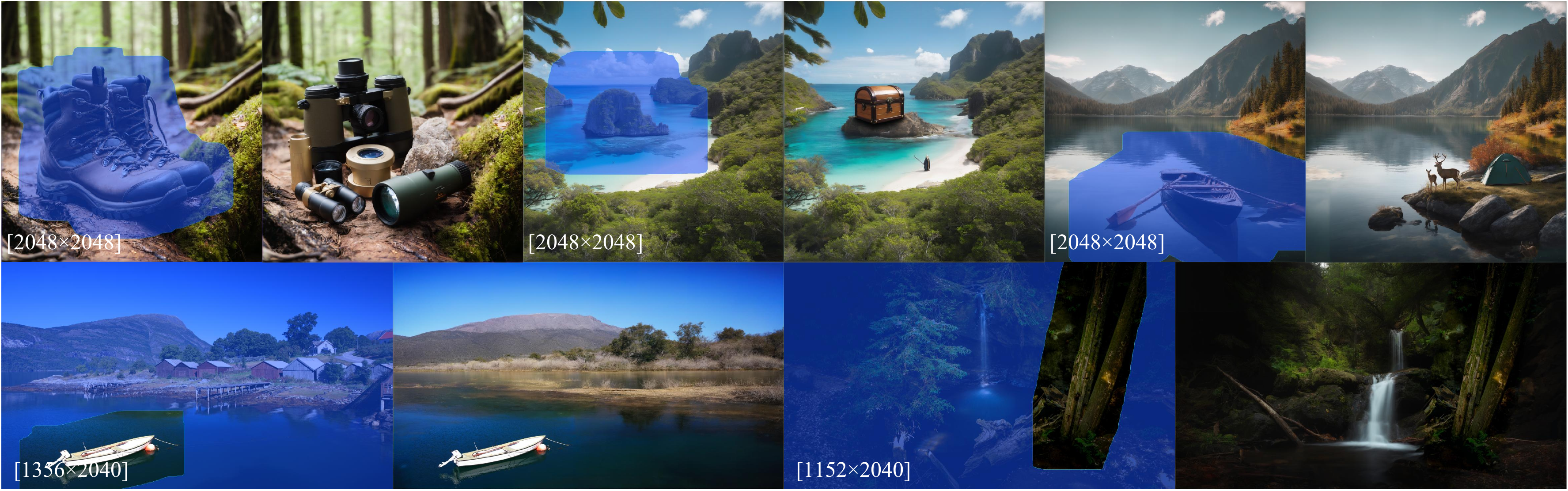}
    \caption{
    {Visual results of image multi-object editing (top) and outpainting (bottom) of ours. The masked regions are in blue.}
    } 
    \label{fig:fig_vis_multiobj}
\end{figure*}

\begin{figure*}[!th]
    \centering
    \includegraphics[width=1\textwidth]{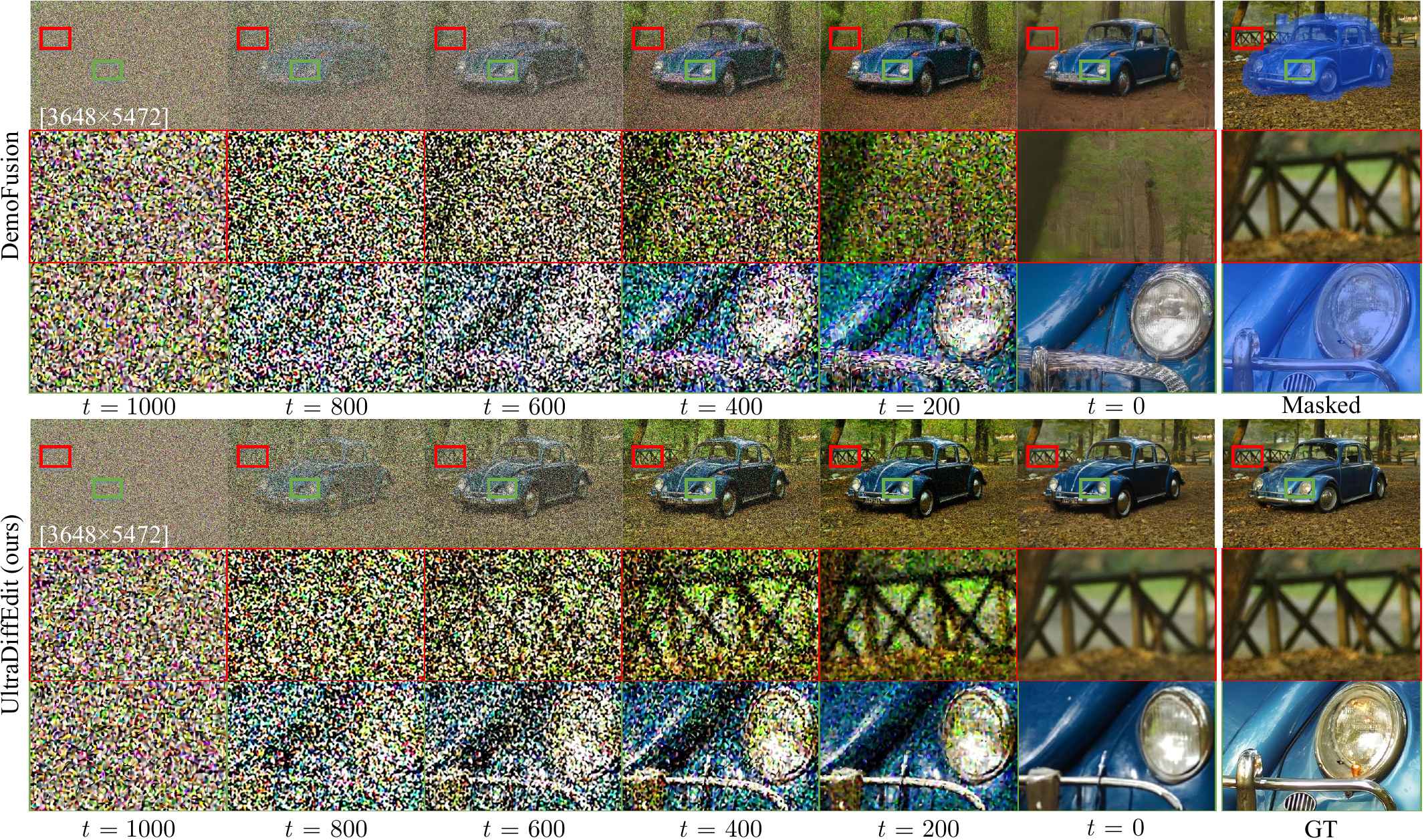}
    \caption{
    Comparison of denoising processes in DemoFusion (top) and UltraDiffEdit (bottom). Compared to DemoFusion, UltraDiffEdit achieves structure refinement at earlier denoising steps, while preserving background information effectively during later steps. Our approach demonstrates superior global-local consistency and produces faithful, detailed results. The masked and ground-truth images are shown in the right column.
    } 
    \label{fig:fig_supp_vis_denoise}
\end{figure*}

\textbf{Extension for existing editing methods.} 
Fig.~\ref{fig:fig_vis_app_control} illustrates our method's extension for existing image editing methods (ControlNet~\cite{zhang2023adding} and IP-adapter~\cite{ye2023ip}) to support high-resolution images.
For ControlNet, we utilized pose keypoints as the conditional input for image editing. 
We first used SDXL-based ControlNet to produce initial edits and then refined these using UltraDiffEdit to achieve high-resolution outputs. 
We utilized IP-adapter~\cite{ye2023ip} to generate visual prompts and plugged into UltraDiffEdit for exemplar-guided~\cite{lu2022inpainting} image editing.
Our UltraDiffEdit enhances detail and fidelity in generated content, significantly extending the capability of existing diffusion models to handle ultra-high-resolution image editing using additional conditioning inputs.
More results can be found in our supplementary document.

{\textbf{Multi-object editing and image outpainting.}
Fig.~\ref{fig:fig_vis_multiobj} (top) illustrates the multi-object editing results using prompts.
We apply free-form masks to local regions and input natural language prompts that describe multiple objects to be included.
UltraDiffEdit inserts specified objects and ensures correct spatial arrangement, proportion, and style consistency within the edited regions.
Fig.~\ref{fig:fig_vis_multiobj} (bottom) showcases the image outpainting results produced by UltraDiffEdit.
We removed background areas and applied binary masks to guide the generation process.
The generated content aligns well with the surrounding image context, maintaining consistent lighting, structure, and texture patterns across the boundary of the edited regions.
}

\textbf{Visualization of denoising processes.}
Fig.~\ref{fig:fig_supp_vis_denoise} illustrates the denoising processes of DemoFusion and our UltraDiffEdit. 
Both models synthesize semantics during early denoising steps and refine image details in later stages. 
Compared to DemoFusion, UltraDiffEdit reveals more structural information at earlier steps while preserving background details within the latent space.
Our multi-patch encoding efficiently embeds large-scale input images into latent space, ensuring both edited and unedited information is retained. Additionally, the global-local consistency denoising generates content and maintains smooth transition at editing boundaries.

\begin{figure*}[t]
    \centering
    \includegraphics[width=1\textwidth]{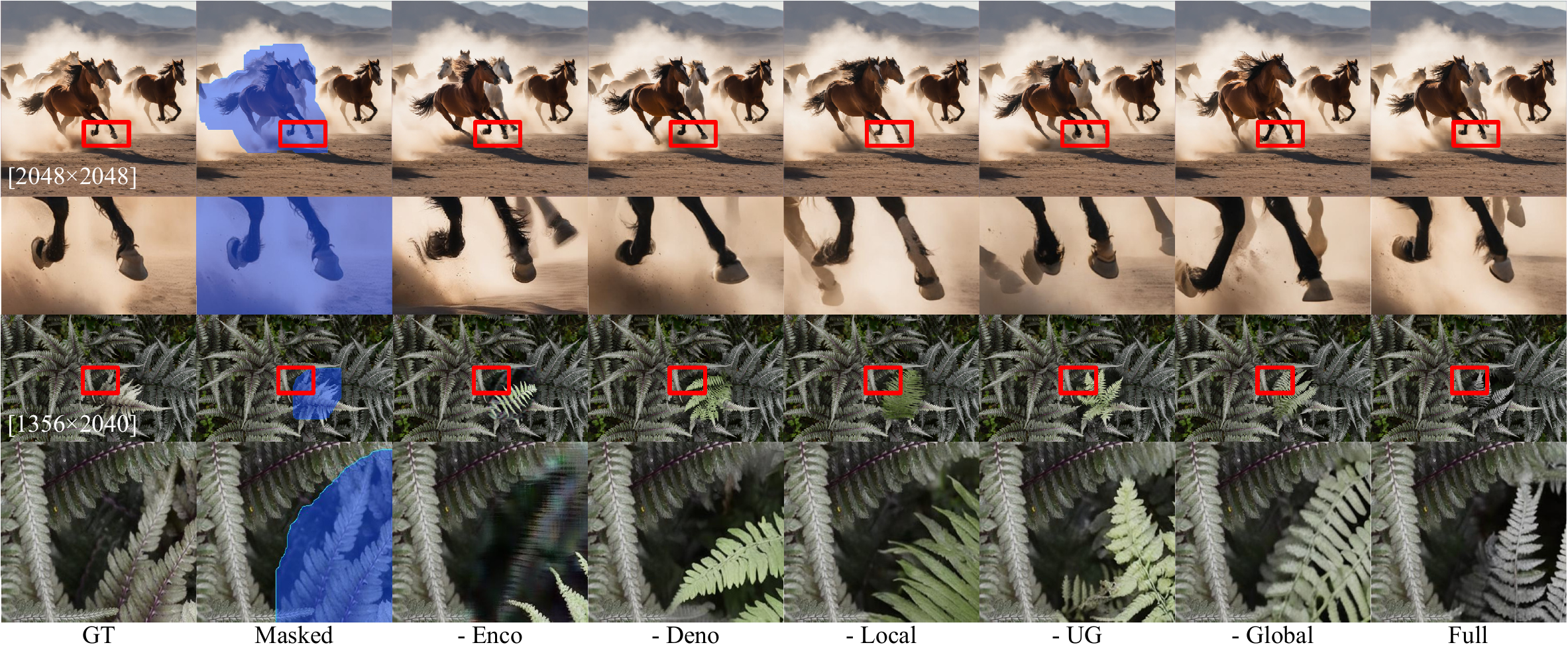}
    \caption{
    Ablation study for the applied components. ``- Enco'' replaces the multi-patch encoding with the original encoding from SDXL; ``- Deno'' substitutes the global-local consistency denoising with the skip residual denoising from DemoFusion;
    ``- Local'' omits the local sampling; ``- UG'' removes the upsampling guidance; ``- Global'' deactivates the dilated global sampling. ``Full'' represents our complete model. 
    Our methods show better global-local consistency with faithful details.
    } 
    \label{fig:fig_vis_ablation}
\end{figure*}

\begin{table}[t]
\centering
\caption{
{User study comparisons between UltraDiffEdit (Ours) and the compared methods. 
We report the percentage of votes, 95\% confidence intervals (CI) of user preference for UltraDiffEdit, $p$-values from two-sided binomial tests, {Chi-Square ($\chi^2$), effect size (Cohen’s $w$), and odds ratio (for ours)}.
}
}
\label{tab:user_study_ablation}
\resizebox{\linewidth}{!}{
\begin{tabular}{cccc}
\hline
{Comparison} & {Percentage of votes} & {95\% CI (Preference)} & {$p$-value} \\
\hline
Ours vs. CoordFill   & {94.52\% vs. 5.48\%} & {[0.924, 0.960]} & {$1 \times 10^{-130}$} \\
Ours vs. HD-Painter   & {70.00\% vs. 30.00\%} & {[0.663, 0.735]} & {$1 \times 10^{-23}$} \\
Ours vs. DemoFusion  & {57.26\% vs. 42.74\%} & {[0.533, 0.611]} & {$3.4 \times 10^{-4}$} \\
Ours vs. SDXL+Inf-DiT & {53.06\% vs. 46.94\%} & [0.491, 0.570] & 0.137 \\
\hline
\hline
{Comparison} & {Chi-Square ($p$-value)} & {Effect size} & {Odds ratio} \\
\hline
{Ours vs. CoordFill}    & {491.5 ($1 \times 10^{-5}$)} & {0.890} & {17.24$\times$} \\
{Ours vs. HD-Painter}  & {99.2 ($1 \times 10^{-4}$)} & {0.400} & {2.33$\times$} \\
{Ours vs. DemoFusion} & {13.1 ($1 \times 10^{-3}$)}  & {0.145} & {1.34$\times$} \\
{Ours vs. SDXL+Inf-DiT} & {2.3 ($1 \times 10^{-1}$)}     & {0.061} & {1.13$\times$} \\
\hline

\end{tabular}
}
\end{table}

\textbf{User study.}
We conducted a user study to perform method-to-method comparisons against CoordFill, HD-Painter, DemoFusion, and SDXL+Inf-DiT, respectively. 
We randomly selected 20 images (questions) from DIV2KEdit for each participant, and each question included a masked image, a ground truth image, text prompts, and shuffled edited images. 
Participants ranked the methods based on visual quality, realism, and text prompt consistency. 
{We recruited 31 participants from both technical and non-technical backgrounds, including researchers, graduate students, undergraduate students, and general users. We ensured that all participants were familiar with basic visual comparison tasks and could reliably assess perceptual image quality differences.}
A total of 620 votes were collected for each method-to-method comparison.

{Table~\ref{tab:user_study_ablation} {(top)} presents the percentage of votes, 95\% confidence intervals (CI), and $p$-values from binomial tests. 
UltraDiffEdit received the majority of votes across all comparisons, significantly outperforming CoordFill (94.52\% vs. 5.48\%), HD-Painter (70.00\% vs. 30.00\%), DemoFusion (57.26\% vs. 42.74\%), and SDXL+Inf-DiT (53.06\% vs. 46.94\%). 
Compared to CoordFill and HD-Painter, UltraDiffEdit shows statistically significant improvements ($p < 10^{-23}$), with markedly higher preference proportions (CI: $[0.924, 0.960]$ and $[0.663, 0.735]$, respectively). 
Similarly, the comparison with DemoFusion yields a significant difference ($p = 3.4 \times 10^{-4}$), reinforcing the perceptual superiority of our approach.
While UltraDiffEdit slightly outperforms SDXL+Inf-DiT in user preference, the difference is not statistically significant ($p = 0.137$), indicating that both methods offer comparable perceptual quality.
{To provide a more comprehensive analysis of these preferences, we further report three complementary statistical metrics (Chi-Square ($\chi^2$), effect size (Cohen’s $w$), and odds ratio) to further validate the statistical significance, as shown in Table~\ref{tab:user_study_ablation} (bottom). The Chi-Square tests show high statistical significance against CoordFill ($\chi^2=491.5$ with $p = 1 \times 10^{-5}$) and HD-Painter ($\chi^2=99.2$ with $p = 1 \times 10^{-4}$), and a significant result against DemoFusion ($\chi^2=13.1$ with $p = 1 \times 10^{-3}$). The effect sizes for these comparisons indicate that the observed preferences are significant.
The odds ratios also show that users prefer our method to CoordFill and HD-Painter. 
Our user study involves a relatively small sample size (31 participants, 20 images each, totaling 620 votes per comparison).
Nevertheless, the user study confirms that UltraDiffEdit achieves consistently superior or competitive performance, with user preference backed by statistical evidence.}

\begin{figure*}[t] 
	\centering
	\includegraphics[width=.96\textwidth]{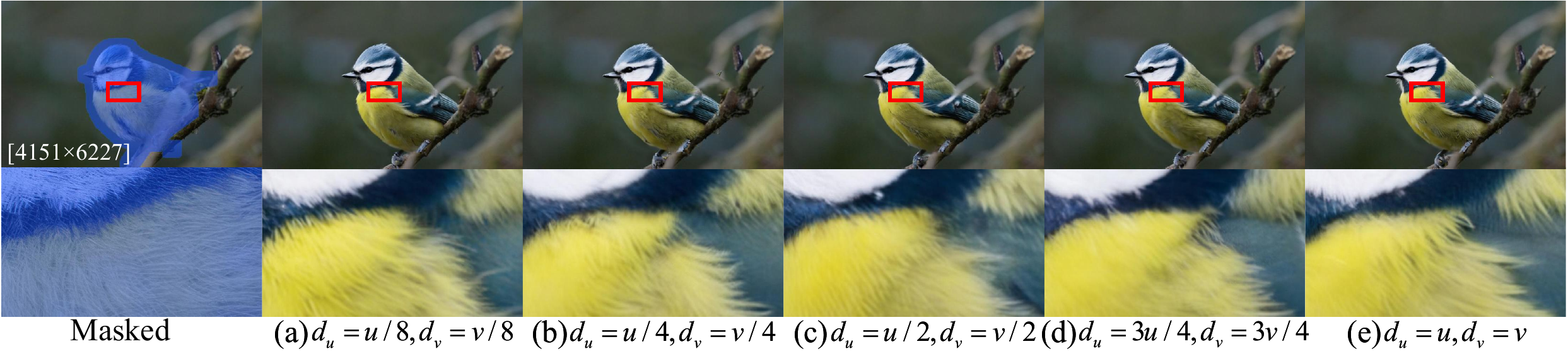}
	\caption{{Visual results using different strides  $d_u$ and $d_v$. Time cost for each configuration: (a) $10198.4$ seconds, (b) $9727.5$ seconds, (c) $9606.7$ seconds, (d) $9580.2$ seconds, (e) $9561.6$ seconds. }}\label{fig:fig_stride_ablation} 
\end{figure*}

\subsection{Ablation study}\label{sec:ablation_study}
To assess the contribution of each component in our framework, we conducted ablation studies by selectively removing individual elements and evaluating their impacts. More visual results are in our supplementary document.

\textbf{Effects of individual components.} As shown in Fig.~\ref{fig:fig_vis_ablation}, replacing the multi-patch encoding (- Enco) with the resizing and encoding method from SDXL results in inadequate preservation of unedited regions in the latent space. 
Substituting the global-local consistency denoising (- Deno) with the skip residual denoising from DemoFusion disrupts the natural transition between masked and unedited regions.   
Removing the local sampling (- Local) leads to reduced detail quality in the output. 
Excluding upsampling guidance (- UG) causes a noticeable decline in both detail and structural consistency. 
Furthermore, without global sampling (- GL), the edited image shows several repeated objects.
In contrast, our full model leverages multi-patch encoding to preserve unedited image details and uses a hybrid sampling approach to incorporate local, upsampling guidance, and global sampling to enhance the richness of information during denoising. This results in better global-local consistency with finer details.

\begin{table}[t]
    \centering
    \caption{Ablation study on the DIV2KEdit and Syn2KEdit datasets. ``- Enco'' replaces the multi-patch encoding with the original encoding from SDXL; ``- Deno'' substitutes the global-local consistency denoising with the skip residual denoising from DemoFusion;
    ``- Local'' omits the local sampling; ``- UG'' removes the upsampling guidance; ``- Global'' deactivates the dilated global sampling. ``Full'' represents our complete model.}
    \label{tab:ablation_study}
    \resizebox{\linewidth}{!}{%
        \begin{tabular}{c|cccc|cccc}
            \hline
            \multirow{1}{*}{} &\multicolumn{4}{c|}{DIV2KEdit (2K)} & \multicolumn{4}{c}{Syn2KEdit (2K)}  \\
            \hline
            \multirow{1}{*}{Config.} &FID$^{\downarrow}$& LPIPS$^{\downarrow}$ &FID$_{\text{crop}}^{\downarrow}$& LPIPS$_{\text{crop}}^{\downarrow}$ &FID$^{\downarrow}$& LPIPS$^{\downarrow}$ &FID$_{\text{crop}}^{\downarrow}$& LPIPS$_{\text{crop}}^{\downarrow}$ \\
            \hline
            - Enco & 84.25 & 0.2215 & 59.47 &  0.2754  &    55.55 &  0.1072  &  30.42 & 0.1551  \\
            - Deno & 85.37   &  0.2187  & 58.31   &   0.2707  & 54.88   &  0.1071    & 30.93  & 0.1540 \\
            - Local & 84.33  & 0.2136&  \textbf{57.65} & 0.2662   &    53.56 & 0.1059 &  30.65 & 0.1518  \\
            - UG & 82.69 & 0.2130 & 59.14 & 0.2657 &   56.57 &  0.1069   & 30.21 &  0.1526 \\
            - Global & 86.97 & 0.2153 &  59.24 & 0.2692  &  54.95 &  0.1070   & 30.71 &  0.1528 \\
            Full &  \textbf{78.59}  &  \textbf{0.2104} &   \textbf{57.65} & \textbf{0.2645}   &      \textbf{52.66} &  \textbf{0.1053} & \textbf{30.05} &\textbf{0.1512}   \\ 
            \hline
        \end{tabular}
    }
\end{table}

Table~\ref{tab:ablation_study} presents the quantitative results of our ablation study on the DIV2KEdit and Syn2KEdit datasets. 
Removing individual components from our full model leads to declines in performance across various metrics, demonstrating the importance of each component.
Replacing the multi-patch encoding (- Enco) with the vanilla encoding method used in SDXL results in significant drops in both global and local perceptual-level metrics. 
This is likely due to the inability to effectively preserve both global and local information in the target image. 
Replacing global-local consistency denoising (- Deno) with skip residual denoising also leads to a drop in quantitative scores. 
This is due to the random noise introduced during the denoising process, which inevitably alters the unedited visual features.
Omitting local sampling (- Local) reduces perceptual scores to some extent, while removing the upsampling guidance (- UG) sampling leads to notable declines in crop-based metrics (e.g., LPIPS$_{\text{crop}}$ and FID$_{\text{crop}}$). 
Disabling global sampling (- GL) affects global perceptual similarity metrics, such as LPIPS and FID.
Our framework, through multi-patch encoding, efficiently preserves high-resolution visual features while avoiding out-of-memory (OOM) issues. 
Additionally, patch-based hybrid sampling, which processes the subsets of feature maps, reduces GPU memory demands and captures visual features at local, intermediate, and global levels, significantly enhancing denoising quality.

{\textbf{Effects of strides $d_u$ and $d_v$ in multi-patch encoding.}
Fig.~\ref{fig:fig_stride_ablation} illustrates the effects and corresponding inference times of using different strides ($d_u, d_v$) in our multi-patch encoding method. 
In general, smaller strides ($d_u$, $d_v$) produce more seamless and visually coherent results but incur higher computational costs due to increased patch overlap. 
Similar observations can also be found in MultiDiffusion~\cite{bar2023multidiffusion} and DemoFusion~\cite{du2023demofusion}.
When smaller strides are used, such as $d_u = u/8$ and $d_v = v/8$, the edited images exhibit finer details. 
In contrast, using larger strides, e.g., $d_u = u$ and $d_v = v$ (i.e., no patch overlap), still maintains good global semantic consistency.
Overall, all configurations result in high-quality outputs without noticeable seams. 
To balance quality and efficiency, we adopt $d_u = u/2$ and $d_v = v/2$ as our default setting.}

\begin{figure}[t]
    \centering
    \includegraphics[width=0.485\textwidth]{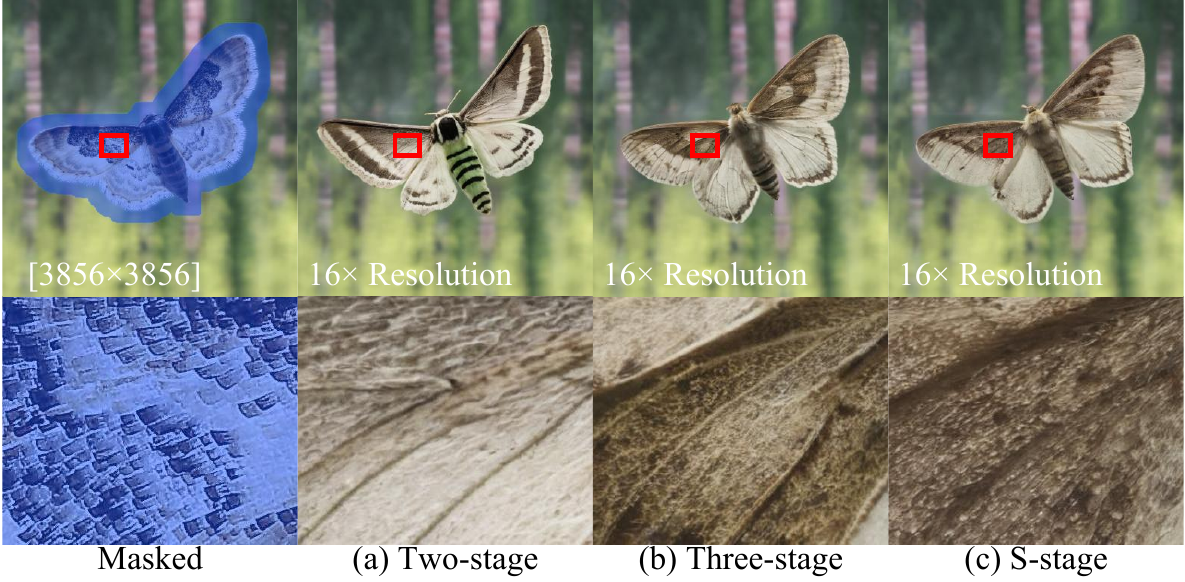}
    \caption{
    Visualization of our UltraDiffEdit using various phase sets: (a) $\mathcal{G}=\{1,S\}$ with two stages, (b) $\mathcal{G}=\{1, \lfloor\frac{S}{2}\rfloor+1, S\}$ with three stages, and (c) $\mathcal{G}=\{1,\ldots,s,\ldots,S\}$ with $S$ stages. Here, the padded image is with $4096 \times 4096$, thus the $S=4=\frac{4096}{1024}$. Time cost for each configuration: (a) $2748$ seconds, (b) $3886$ seconds, (c) $4178$ seconds.
    } 
    \label{fig:fig_supp_stage_ana}
\end{figure}

\textbf{Effects of editing stages in the phase set $\mathcal{G}$.} Fig.~\ref{fig:fig_supp_stage_ana} illustrates the visual results of various phase sets. 
We tested three variants with different configurations for multi-scale progressive editing: (a) $\mathcal{G}=\{1,S\}$ with two stages, (b) $\mathcal{G}=\{1, \lfloor\frac{S}{2}\rfloor+1, S\}$ with three stages, and (c) $\mathcal{G}=\{1,\ldots,s,\ldots,S\}$ with $S$ stages. 
The results indicate that using more stages increases the editing time. 
The two-stage configuration $\mathcal{G}=\{1,S\}$ achieves results similar to the three-stage setup $\mathcal{G}=\{1, \lfloor\frac{S}{2}\rfloor+1, S\}$ but with reduced editing time. 
Employing more stages yields sharper details in the generated content but needs more time for inference.

\begin{figure}[t]
	\centering
	\includegraphics[width=0.4\textwidth]{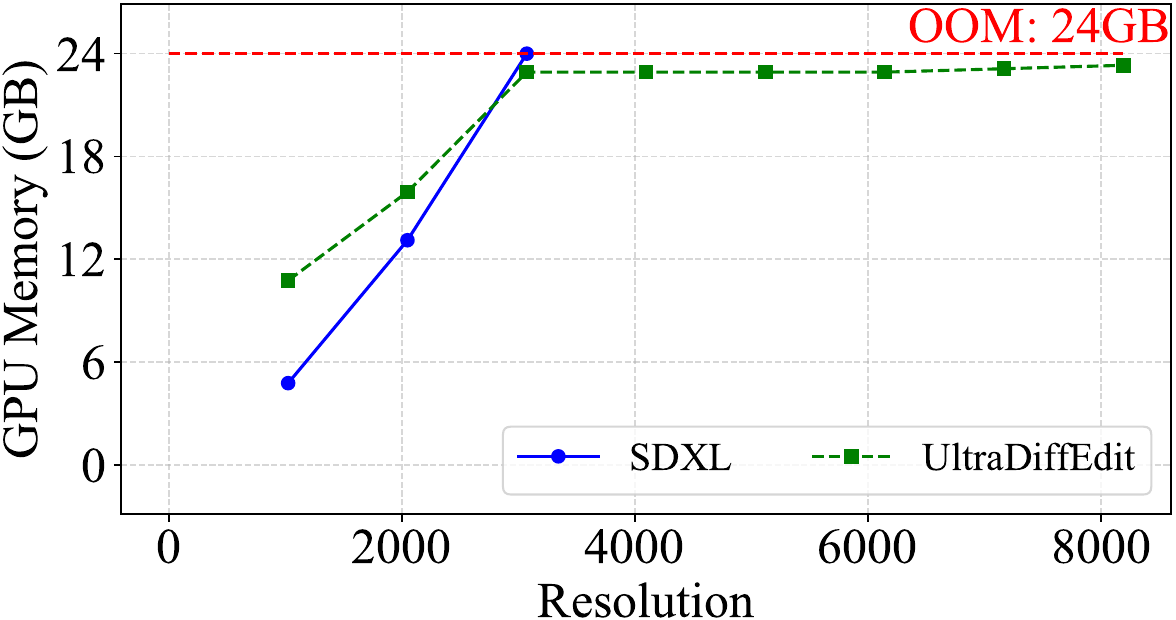}
	\caption{Memory demands of SDXL and UltraDiffEdit, inference on a single image. SDXL encounters out-of-memory issues when processing 3K resolution images.}\label{fig:plot_GPU_memory_UltraDiffEdit} 
\end{figure}
\begin{figure}[t]
	\centering
    \includegraphics[width=0.42\textwidth]{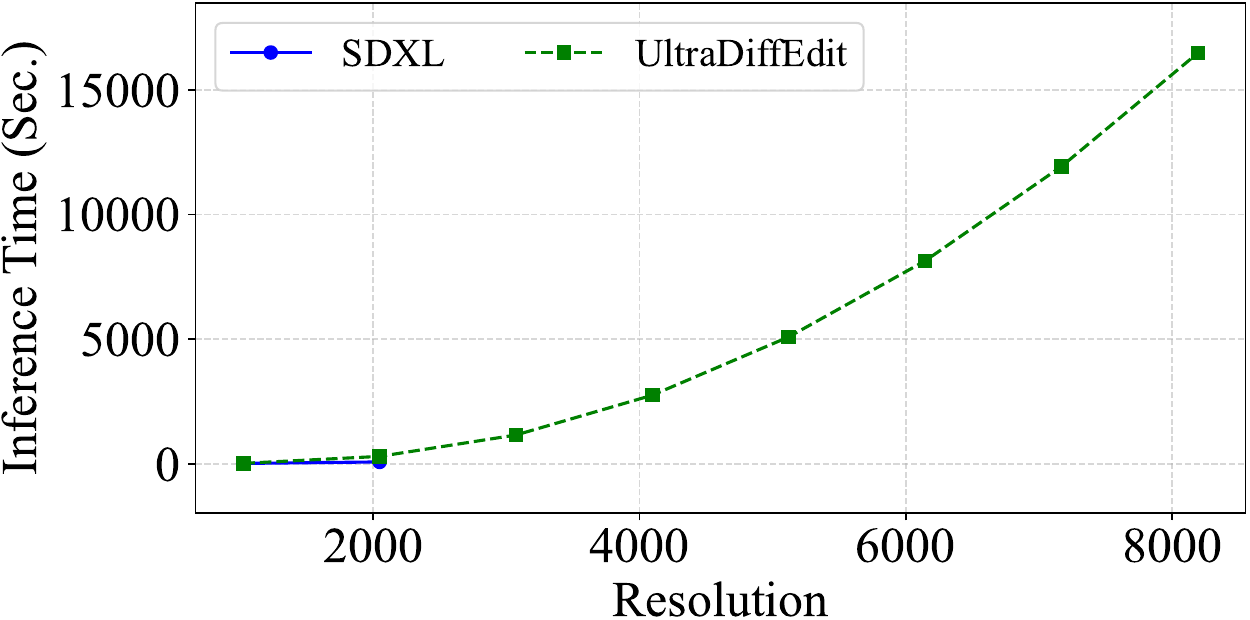}
	\caption{Inference time of SDXL and UltraDiffEdit, inference on a single image. SDXL encounters out-of-memory issues when processing 3K resolution images.}\label{fig:plot_speed_UltraDiffEdit} 
\end{figure}

\subsection{{Discussions}}

\textbf{Memory demands and inference times.}
\revise{Fig.~\ref{fig:plot_GPU_memory_UltraDiffEdit} shows GPU memory demands of UltraDiffEdit (with a mini-batch size of 16 for inference) and SDXL~\cite{podell2023sdxl}. 
When processing larger resolutions (e.g., 3K), SDXL encounters out-of-memory issues. 
In contrast, UltraDiffEdit avoids these memory limitations, albeit with additional computational time.}

\revise{We also show inference time at each scale in Fig.~\ref{fig:plot_speed_UltraDiffEdit}. The inference time scales approximately quadratically with resolution, consistent with our patch-wise design where the number of patches grows with image area. 
UltraDiffEdit remains operable on a single consumer-grade GPU across all tested resolutions up to 8K, whereas SDXL fails with OOM errors beyond 3K.
The increased time cost arises from two main factors. First, our multi-patch encoding method performs patch-based inference to preserve large-scale edited image information in the latent space.
Second, our patch-based hybrid sampling enhances denoising quality by integrating local, intermediate, and global features. 
Despite the extra time, this approach enables high-quality ultra-high-resolution editing.}

UltraDiffEdit prioritizes high-quality outputs on a single moderate GPU (e.g., NVIDIA 3090), making it particularly suited for professional ultra-high-resolution use cases. 
{The term ``ultra'' in our method refers specifically to the ultra-high-resolution capability, not to high computational efficiency.}
{As shown in Table~\ref{tab:gpu_memory_comparison}, we conducted a comparison of inference times and GPU memory usage on an NVIDIA A800 (80GB) for the 2K resolution. 
Although our method is slower, this trade-off guarantees superior image quality. Moreover, users can preview low-resolution intermediate results with our multi-scale progressive editing.
Nevertheless, we are actively exploring multi-GPU inference to enhance efficiency and better balance quality with speed for broader real-world applications.}

\begin{table}[t]
\centering
\caption{{Comparisons in terms of runtime and memory usage for 2K resolution image editing.}}
\resizebox{\linewidth}{!}{%
    \begin{tabular}{c|cc}
    \hline
    \makebox[0.3\columnwidth][c]{Method} & \makebox[0.3\columnwidth][c]{Time (s)} & \makebox[0.3\columnwidth][c]{Memory (MB)}  \\
    \hline
    CoordFill~\cite{CoordFill_aaai} & {3.26} & 924\\
    HD-Painter~\cite{manukyan2023hd} &   43.10 & 13,566\\
    DemoFusion~\cite{du2023demofusion} &  93.30  &  10,860 \\
    SDXL~\cite{podell2023sdxl}+bicubic & 5.66 & 10,717\\
    SDXL+SRGAN~\cite{Ledig_2017_CVPR}  & 7.08 & 10,717\\
    SDXL+BSRGAN~\cite{zhang2021designing}& 9.58 & 10,717\\
    SDXL+Inf-DiT~\cite{InfDiT} & 5.37 &10,717\\
    Flux~\cite{flux2024} & 133.73 & 41,982\\
    PIXART-$\Sigma$~\cite{chen2024pixart} & 17.95 & 22,167\\
    Wuerstchen~\cite{pernias2024wrstchen} & 20.15 & 14,412\\
    Sana~\cite{xie2024sana} & 9.09 & 15,442\\
    {Megafusion~\cite{wu2025megafusion}}  &  {22.35} & {16,465}\\ 
    {Diffusion-4k~\cite{zhang2025diffusion}} & {73.22} & {23,173}\\
    {RectifiedHR~\cite{yang2025rectifiedhr}} & {27.31} & {15,013}\\
    {URAE~\cite{yu2025ultraresolution}} &  {157.27} & {27,431}\\
    UltraDiffEdit & 109.71 &  11,014\\
    \hline
    \end{tabular}}
    \label{tab:gpu_memory_comparison}
\end{table}

\begin{figure}[!t]
     \centering
    \includegraphics[width=0.485\textwidth]{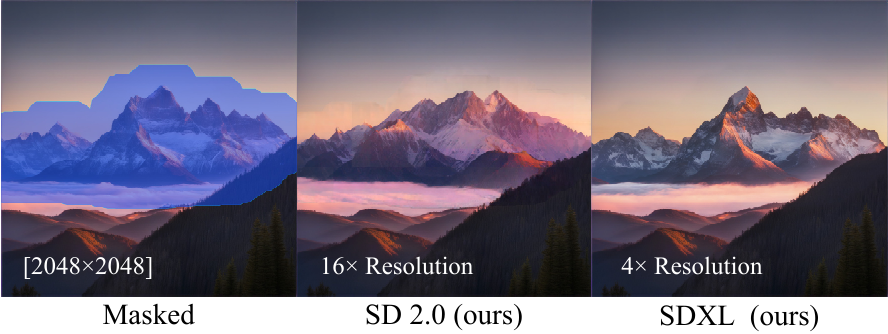}
    \caption{
    Results of UltraDiffEdit on other LDMs, i.e., Stable Diffusion 2.0~\cite{rombach2022high} (default resolution of 512$^2$) and SDXL~\cite{podell2023sdxl} (default resolution of 1024$^2$).
    The ``$x \times$ Resolution'' indicates the scaling factor relative to the original trained resolution.
    }
    \label{fig:fig_vis_SD}
\end{figure}

\begin{figure}[!t]
     \centering
    \includegraphics[width=0.485\textwidth]{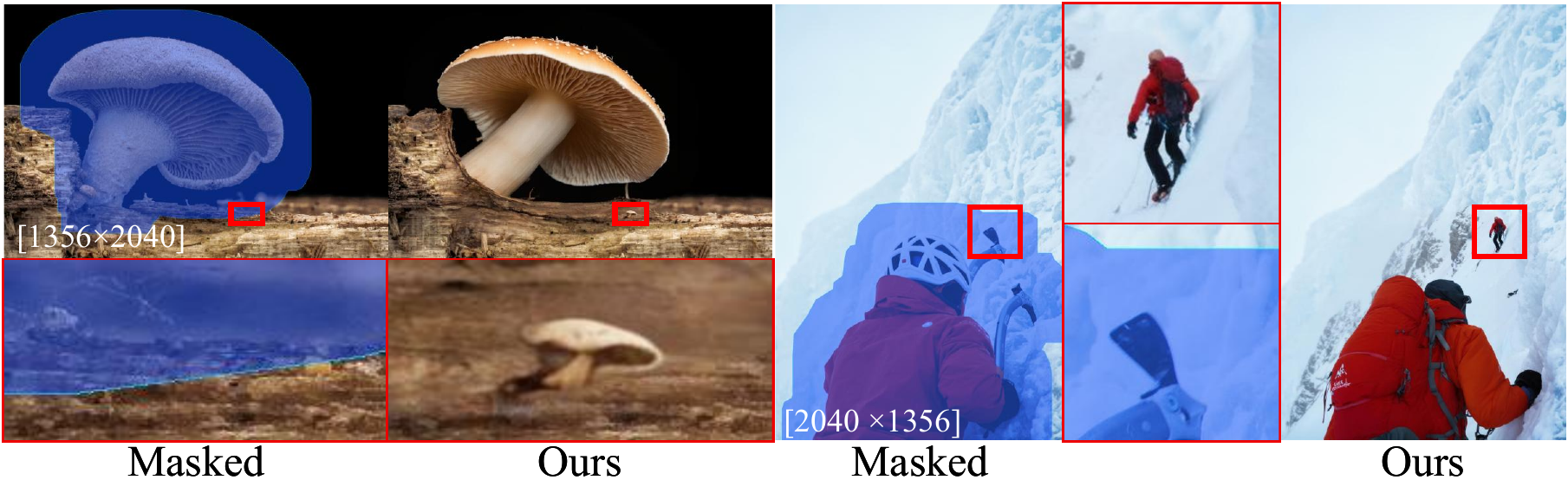}
    \caption{
    {Failure cases of our UltraDiffEdit. Redundant small objects were generated in specific regions.}}
    \label{fig:fig_failure_cases}
\end{figure}

\textbf{Performance with underlying diffusion models.}
The performance of UltraDiffEdit is heavily reliant on the capabilities of the underlying diffusion models.  As shown in Fig.~\ref{fig:fig_vis_SD}, results based on SDXL~\cite{podell2023sdxl} generally outperform those generated using previous versions such as Stable Diffusion 2.0~\cite{rombach2022high}. The performance of UltraDiffEdit can be further improved by training more sophisticated diffusion models.

{\textbf{Failure cases.}
As shown in Fig.~\ref{fig:fig_failure_cases}, redundant small objects were generated in specific regions.
Similar to the observation in DemoFusion~\cite{du2023demofusion}, in patch-based inference, applying text prompts globally across patches may result in the generation of multiple small objects.
A potential solution is to generate multiple candidate images and select the most consistent one.}

\section{Conclusion}
In this paper, we have presented UltraDiffEdit, a novel tuning-free framework for ultra-high-resolution image editing based on pre-trained latent diffusion models. 
It employs {a multi-scale progressive editing strategy that} iteratively integrates generated high-resolution content with unedited areas in a coarse-to-fine manner. 
We developed {a multi-patch encoding mechanism} to preserve unedited visual details in the latent space, as well as a global-local consistency denoising technique to ensure smooth transition between generated and edited regions. 
Furthermore, our {hybrid patch-based sampling approach} captures local, intermediate, and global features, {enhancing both fine detail and semantic coherence during the denoising process. Extensive experiments validate the superior editing quality, robustness, and scalability of UltraDiffEdit across diverse high-resolution editing scenarios}.


\clearpage
\appendices
\section*{Supplementary Document}
\setcounter{figure}{0}
\renewcommand{\thefigure}{A\arabic{figure}}
\setcounter{table}{0}
\renewcommand{\thetable}{A\arabic{table}}
\setcounter{equation}{0}
\renewcommand{\theequation}{A\arabic{equation}}



In this supplement, we describe the implementation of UltraDiffEdit in Section~\ref{sec:implementation_supp}. Section~\ref{sec:data_supp} covers the collection process for our three benchmark datasets for high-resolution image editing. In Section~\ref{sec:exp_supp}, we present further experimental results, including qualitative examples of ultra-high-resolution editing and experiments with additional conditional inputs.

\section{More implementation details}\label{sec:implementation_supp}
\textbf{Implementation of UltraDiffEdit.} 
We utilized Python and PyTorch to build the proposed framework. 
We conducted all the experiments on the NVIDIA GeForce RTX 3090 GPU with 24 GB memory.
The pre-trained models for Stable Diffusion (SD) 1.5 and 2.0~\cite{rombach2022high} (default resolution of 512$^2$) are sourced from ``runwayml/stable-diffusion-inpainting'' and ``stabilityai/stable-diffusion-2-inpainting,'' respectively. 
The pre-trained model for SDXL~\cite{podell2023sdxl} (default resolution of 1024$^2$) is from ``stabilityai/stable-diffusion-xl-base-1.0.''

\textbf{Compared methods.} 
We used the authors' codes for CoordFill~\cite{CoordFill_aaai}, HD-Painter~\cite{manukyan2023hd}, DemoFusion~\cite{du2023demofusion}, SDXL~\cite{podell2023sdxl}, SRGAN~\cite{Ledig_2017_CVPR}, BSRGAN~\cite{zhang2021designing}, and Inf-DiT~\cite{InfDiT}. 
For SDXL~\cite{podell2023sdxl}+bicubic, SDXL+SRGAN~\cite{Ledig_2017_CVPR} SDXL+BSRGAN~\cite{zhang2021designing}, and SDXL+Inf-DiT~\cite{InfDiT}, we downsampled the input image to 1K resolution and upsampled the initial output from SDXL using enhancement models mentioned above for aligning the high-resolution input.
We maintained downsampled input aspect ratios to accommodate the resolution limits of SDXL.

\section{Data collection of benchmark datasets}\label{sec:data_supp}
For high-resolution real-image editing, we developed three benchmark datasets: DIV2KEdit, Syn2KEdit, and UHRSDEdit. 
As shown in Fig.~\ref{fig:fig_img_example}, each dataset includes images, text prompts, editing masks, and additional conditioning inputs (Canny edges, depth maps, and pose keypoints).
Fig.~\ref{fig:fig_resolution} shows the density plots of image resolutions in DIV2KEdit and UHRSDEdit. Fig.~\ref{fig:fig_mask_ratio} presents the mask statistics for the three datasets mentioned above.

\textbf{DIV2KEdit.}
DIV2KEdit images come from the DIV2K~\cite{DIV2K} validation set, containing 100 images with 2K resolution. 
The masks for editing were generated using the semantic segmentation tool X-AnyLabeling~\cite{X_AnyLabeling}, with manual adjustments to define the editing regions. We also applied OpenCV's dilation operation (kernel size 100, three iterations) to create masks. 
Text prompts were generated with the BLIP-2~\cite{li2023blip} model.

\textbf{Syn2KEdit.}
Syn2KEdit is a synthetic dataset of 100 images at 2K resolution ($2048 \times 2048$ for each image) featuring diverse styles and objects. We utilized ChatGPT~\cite{ChatGPT} to generate 100 high-quality prompts, which were used with DemoFusion to create images. 
Editing masks were similarly generated using the semantic segmentation tool  X-AnyLabeling~\cite{X_AnyLabeling}, manual adjustments, and OpenCV's dilation operation.

\textbf{UHRSDEdit.}
UHRSDEdit is derived from the UHRSD~\cite{Xie_2022_CVPR} test set, which consists of 988 images and saliency masks at 4K-8K resolution. We generated editing masks by applying OpenCV's dilation operation to the saliency masks (kernel size 100, three iterations). Text prompts for editing were created using the BLIP-2~\cite{li2023blip}.

\textbf{Conditional inputs for ControlNet applications.}
We generated conditional inputs, such as Canny edges, depth maps, and pose keypoints, for all three datasets. High-resolution images were downsampled to 1K for these tasks (keeping the aspect ratio). Canny edges were produced using the Canny algorithm (thresholds: 100, 200). Depth maps were generated using DPT~\cite{Ranftl_2021_ICCV}, and pose skeletons were detected with OpenPose~\cite{OpenPose}.

\begin{figure*}[!ht]
    \centering
    \includegraphics[width=1.0\textwidth]{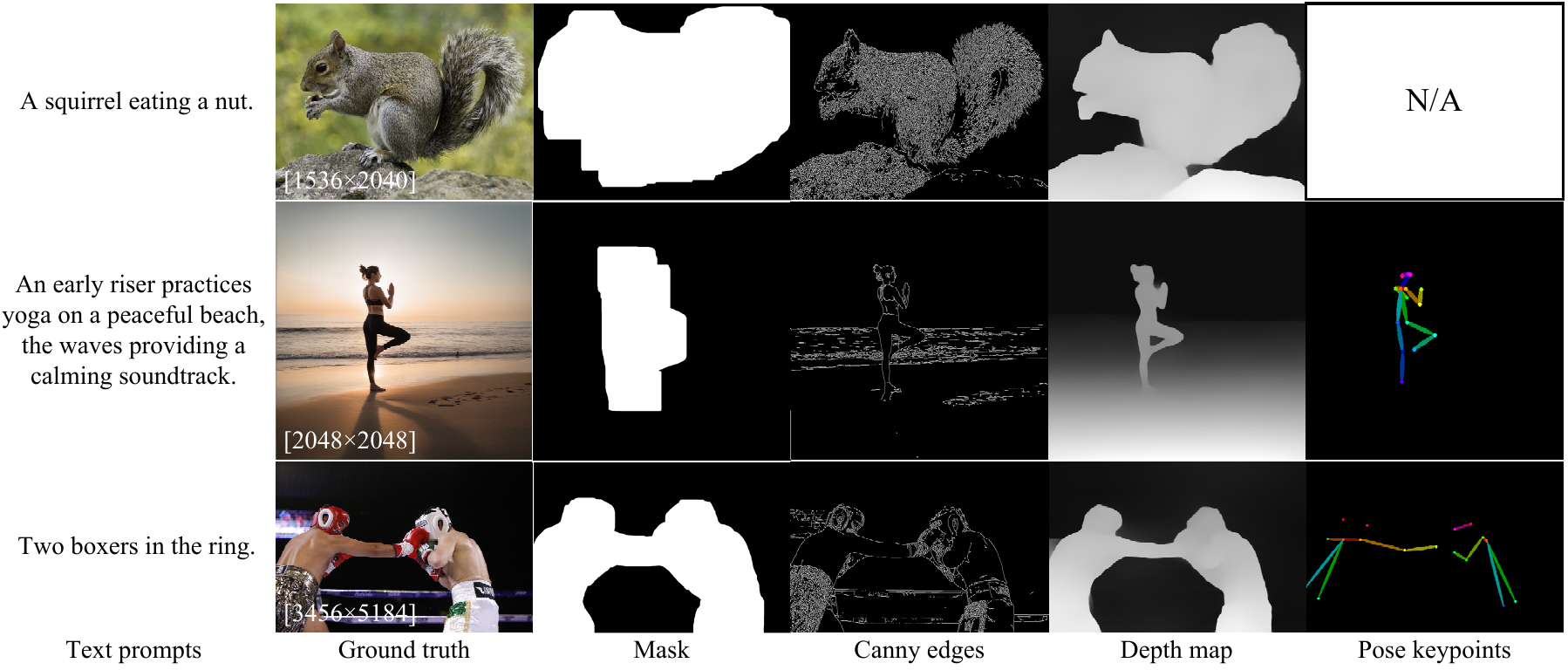}
    \caption{
    Examples of text prompts, ground-truth images, mask images, Canny edges, depth maps, and pose keypoints from the three developed benchmark datasets: DIV2KEdit (row 1), Syn2KEdit (row 2), and UHRSDEdit (row 3).
    ``N/A'' indicates that there are no keypoints for images without human-body objects.
    } 
    \label{fig:fig_img_example}
\end{figure*}

\begin{figure*}[!ht]
    \centering
    \includegraphics[width=0.75\textwidth]{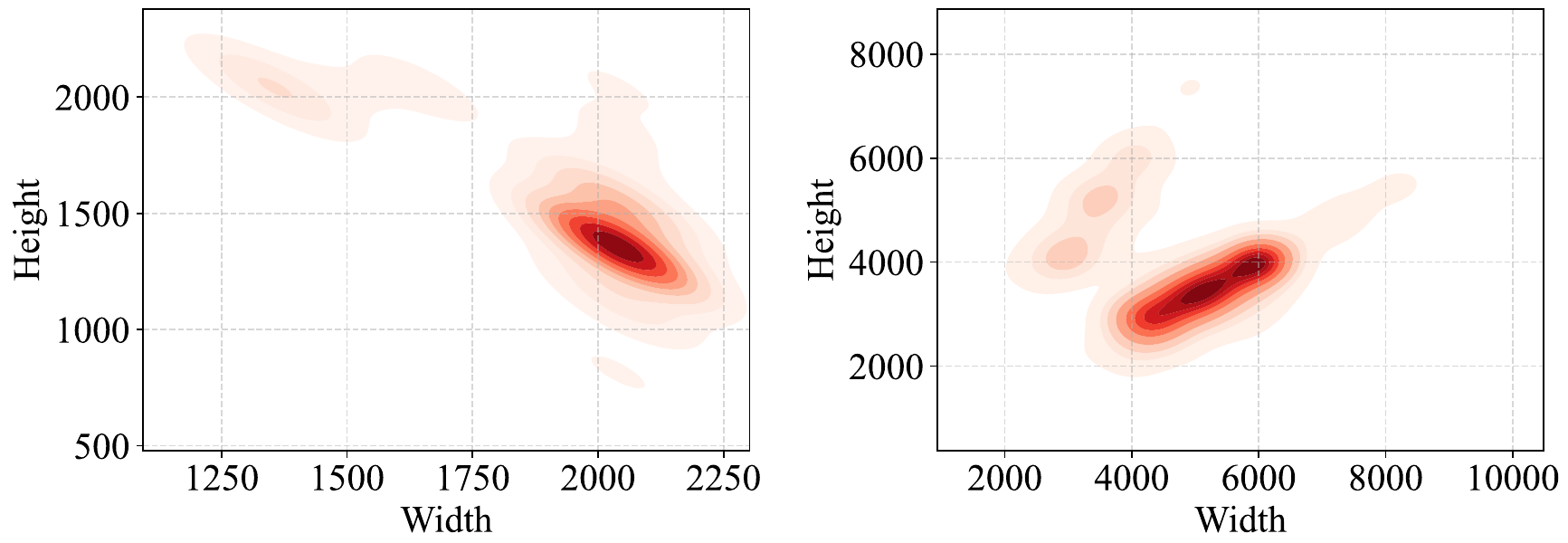}
    \caption{
    Density plots of image resolutions in the developed DIV2KEdit (left) and UHRSDEdit (right).
    } 
    \label{fig:fig_resolution}
\end{figure*}

\begin{figure*}[!ht]
    \centering
    \includegraphics[width=0.9\textwidth]{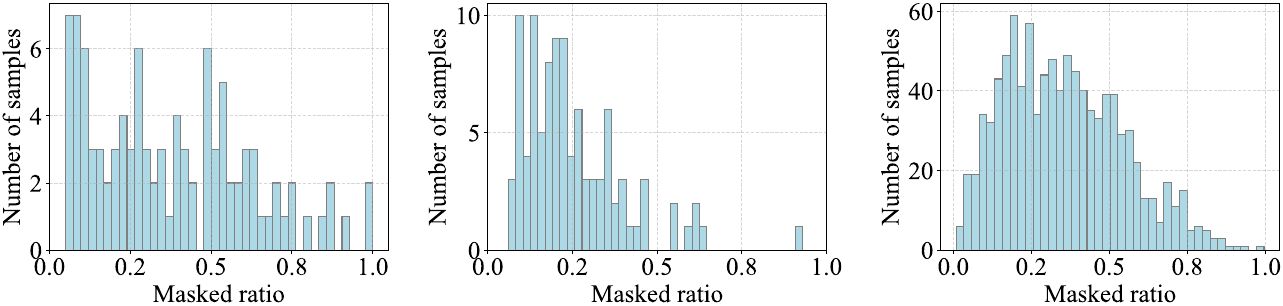}
    \caption{
    Mask statistics of the three developed benchmark datasets: DIV2KEdit (left), Syn2KEdit (middle), and UHRSDEdit (right).
    } 
    \label{fig:fig_mask_ratio}
\end{figure*}

\begin{figure*}[t]
    \centering
    \includegraphics[width=0.96\textwidth]{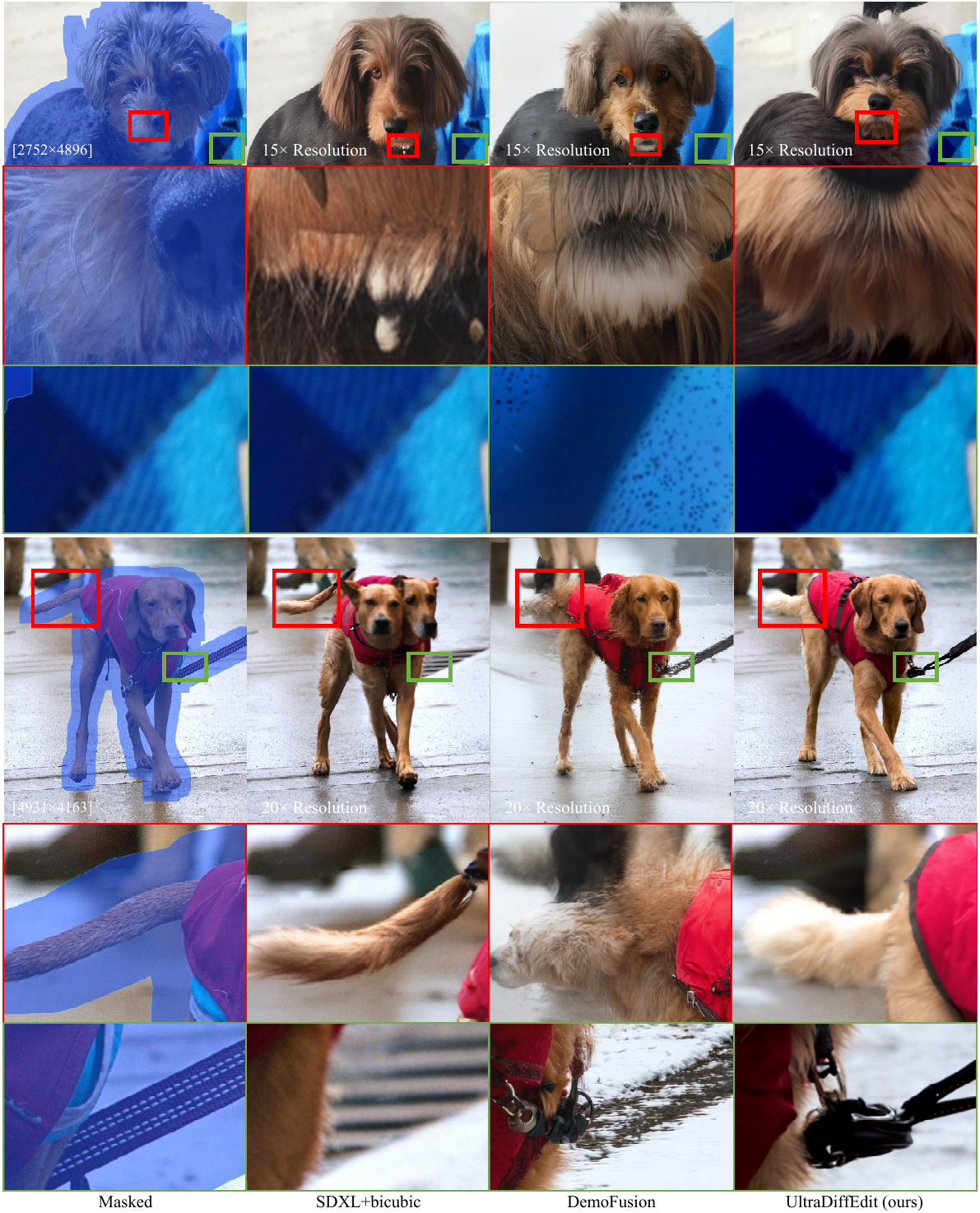}
    \caption{
   More visual comparisons for ultra-high-resolution image editing.
    } 
    \label{fig:fig_vis_supp_HURSD_compare1}
\end{figure*}

\begin{figure*}[t]
    \centering
    \includegraphics[width=0.96\textwidth]{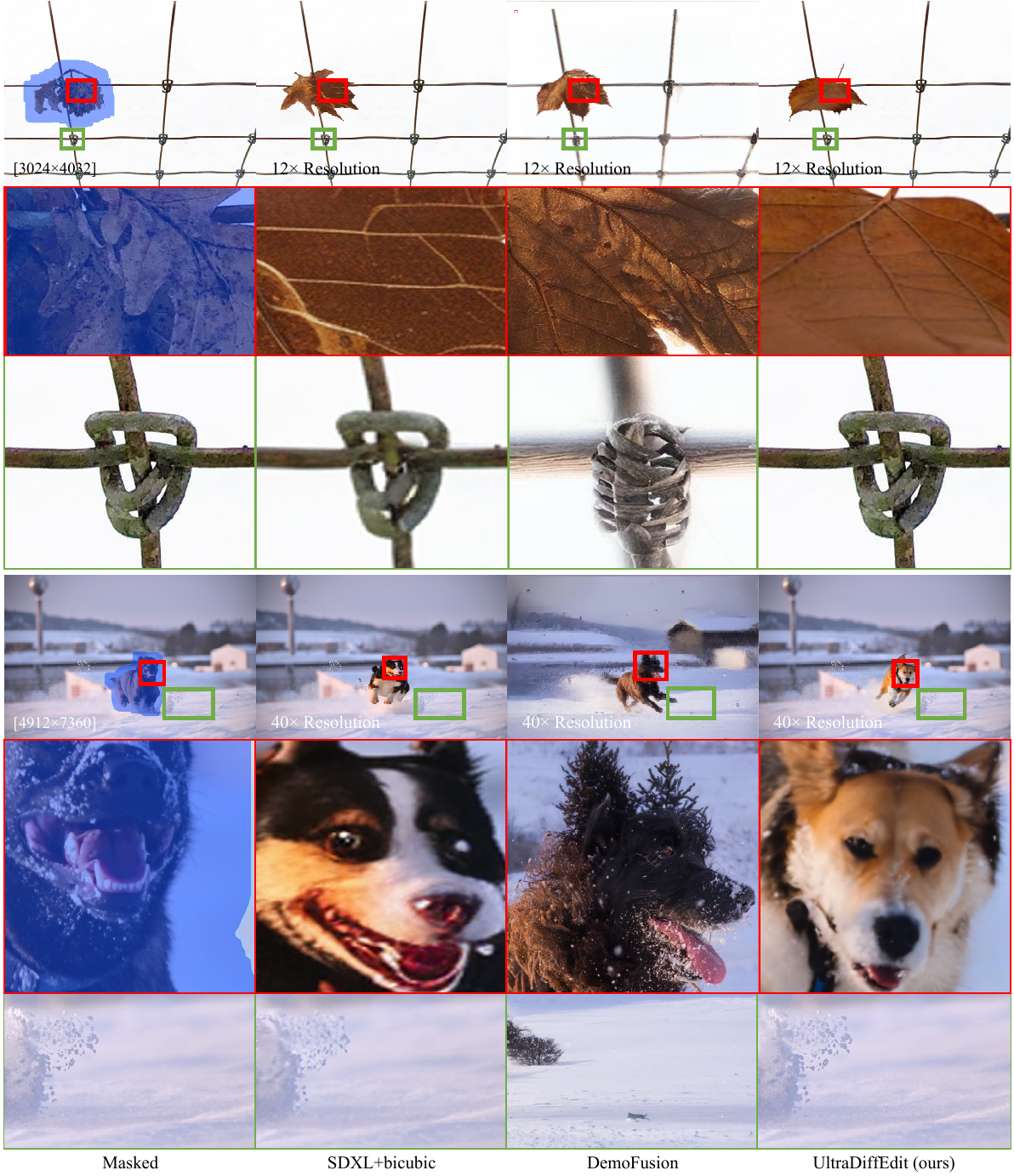}
    \caption{
    More visual comparisons for ultra-high-resolution image editing.
    } 
    \label{fig:fig_vis_supp_HURSD_compare2}
\end{figure*}

\begin{figure*}[t]
    \centering
    \includegraphics[width=0.88\textwidth]{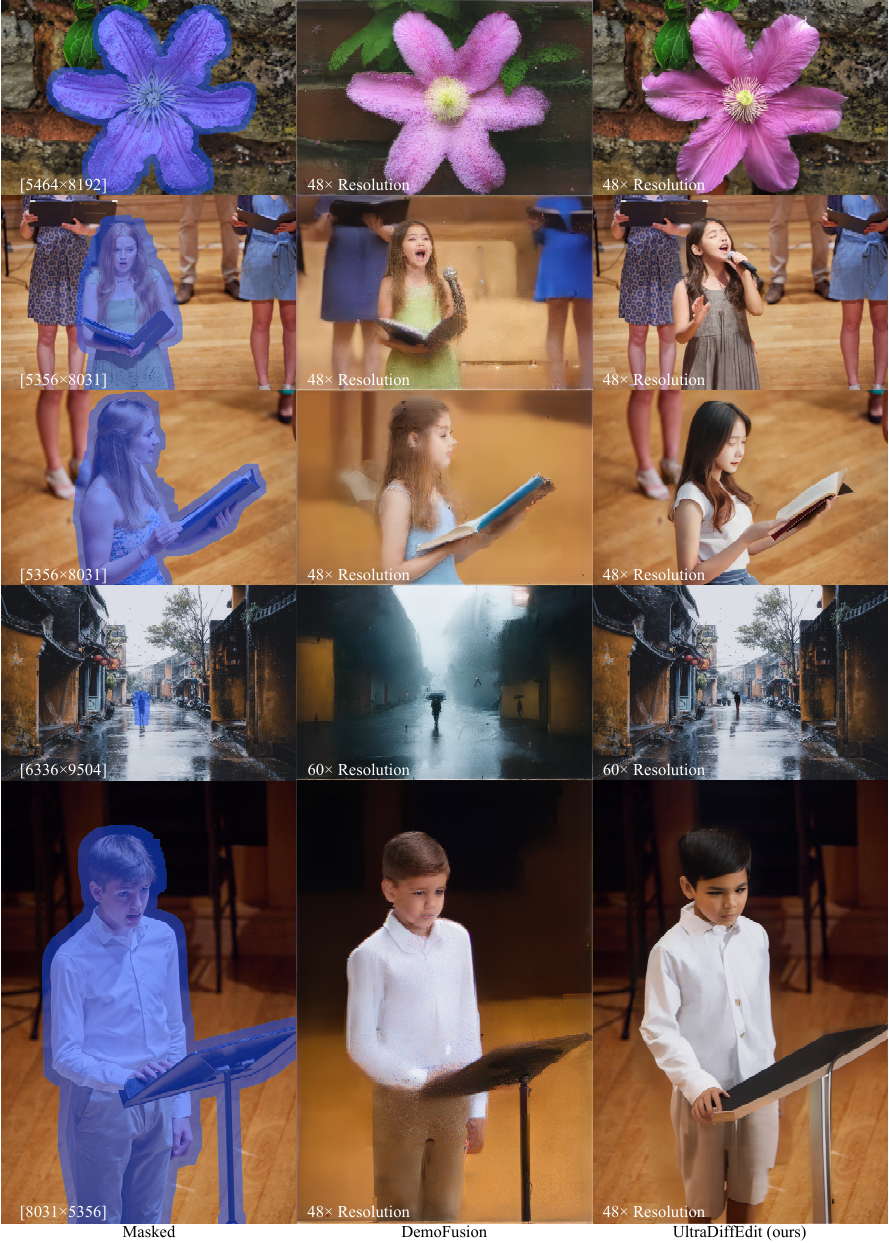}
    \caption{
    More visual comparisons for ultra-high-resolution (8K) image editing.
    } 
    \label{fig:fig_vis_supp_HURSD_8K}
\end{figure*}

\begin{figure*}[t]
    \centering
    \includegraphics[width=0.96\textwidth]{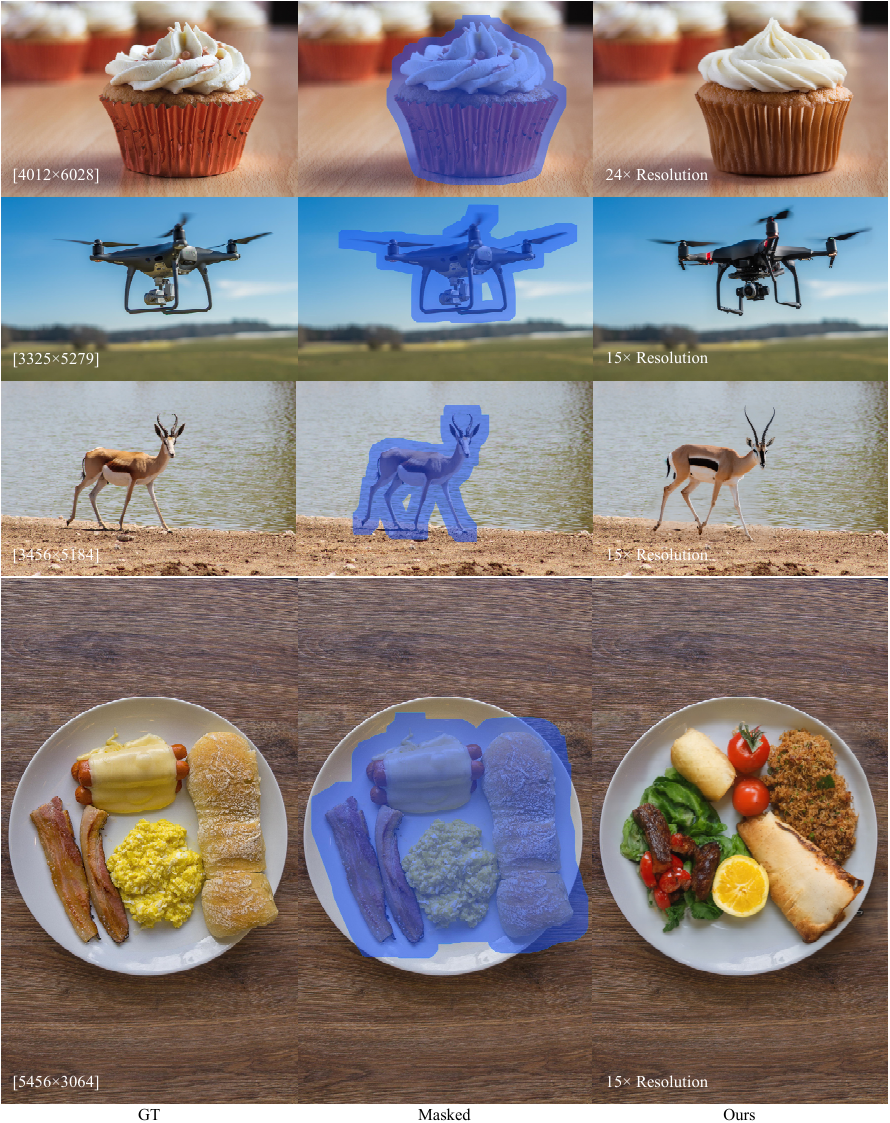}
    \caption{
    Visual results of UltraDiffEdit for ultra-high-resolution image editing.
    } 
    \label{fig:fig_supp_ultra}
\end{figure*}

\begin{figure*}[t]
    \centering
    \includegraphics[width=0.94\textwidth]{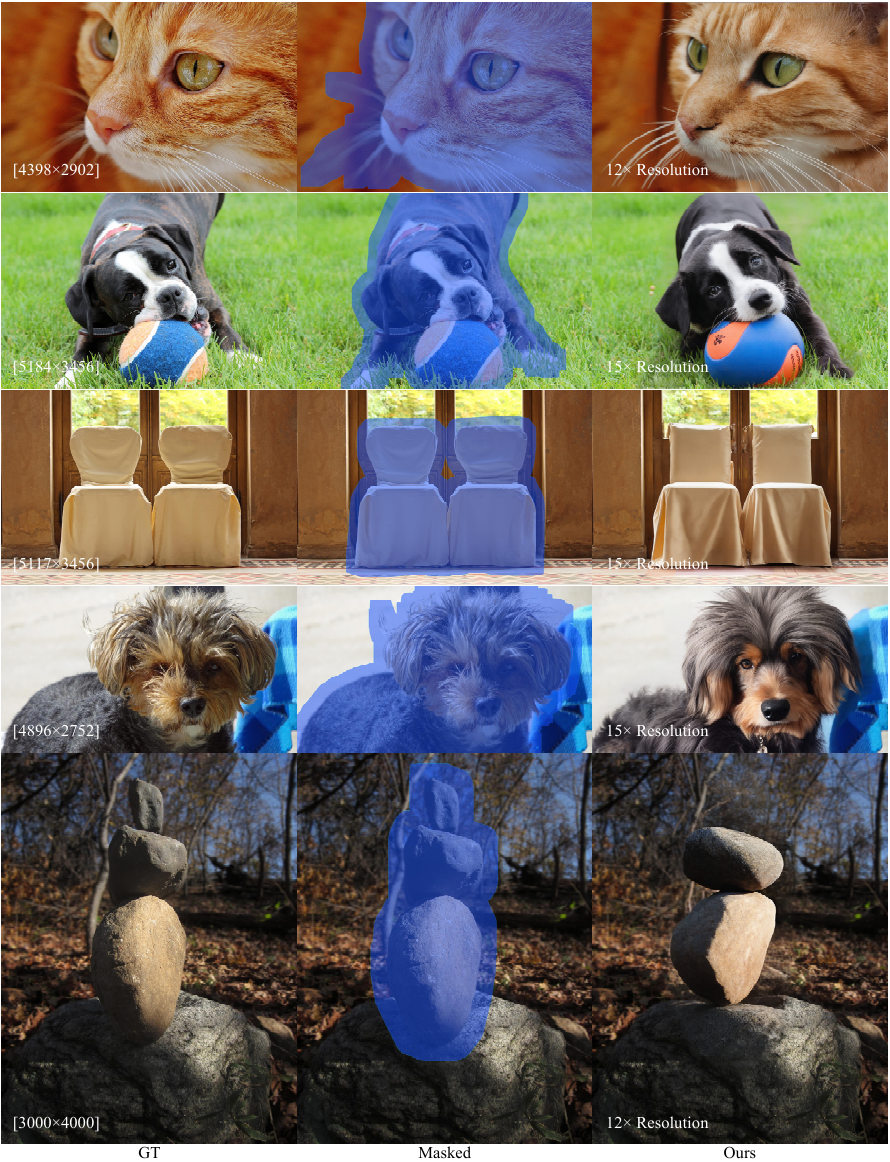}
    \caption{
    Visual results of UltraDiffEdit for ultra-high-resolution image editing.
    } 
    \label{fig:fig_supp_ultra2}
\end{figure*}

\begin{figure*}[t]
    \centering
    \includegraphics[width=0.98\textwidth]{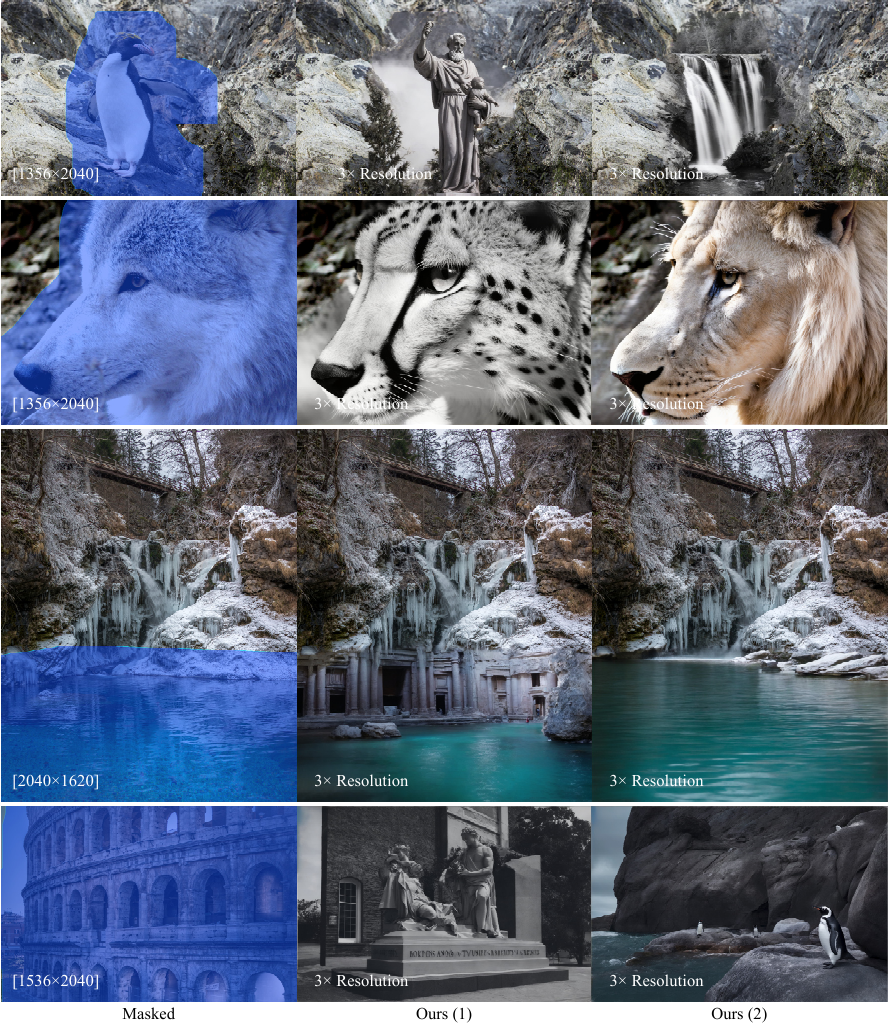}
    \caption{
    Visual results of UltraDiffEdit for high-resolution image editing using various text prompts.
    } 
    \label{fig:fig_supp_text_editing}
\end{figure*}

\begin{figure*}[t]
    \centering
    \includegraphics[width=0.90\textwidth]{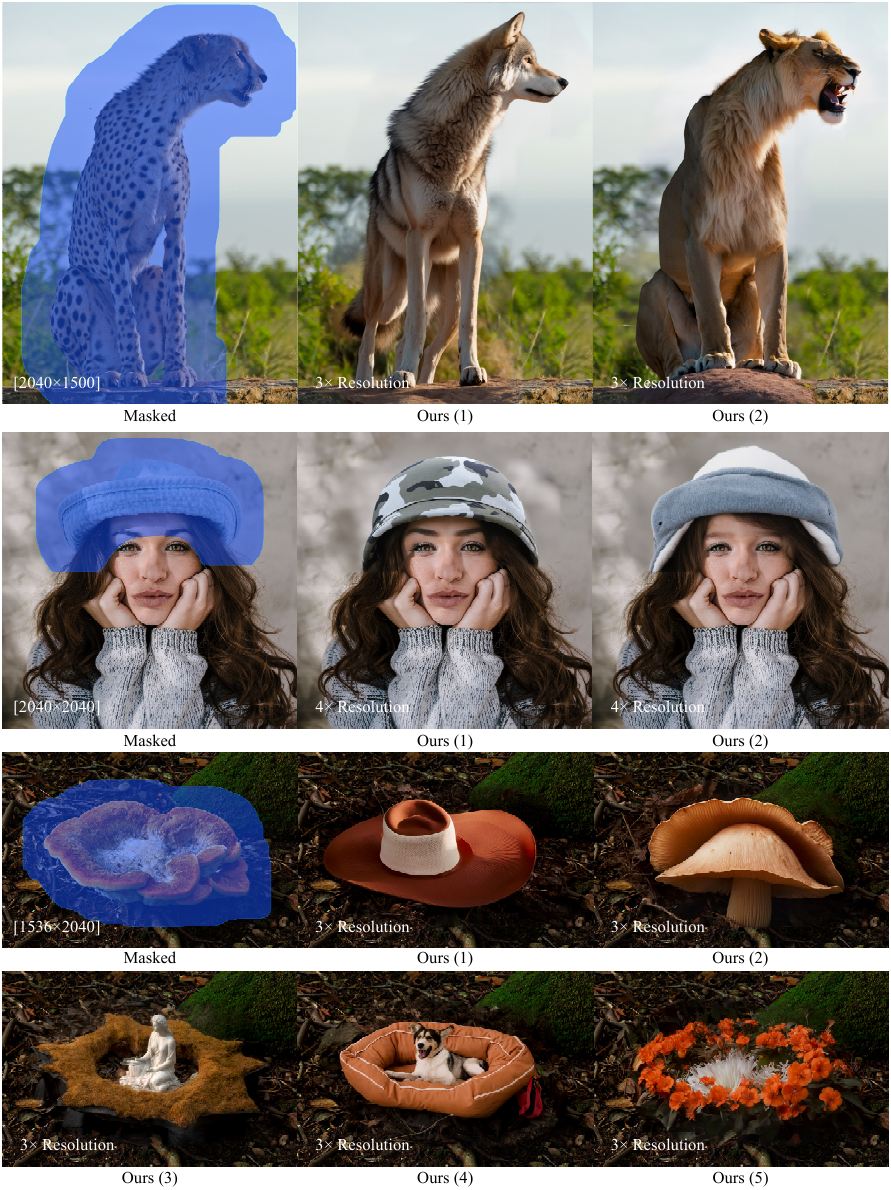}
    \caption{
    Visual results of UltraDiffEdit for high-resolution image editing using various text prompts.
    } 
    \label{fig:fig_supp_text_editing2}
\end{figure*}

\begin{figure*}[t]
    \centering
    \includegraphics[width=0.94\textwidth]{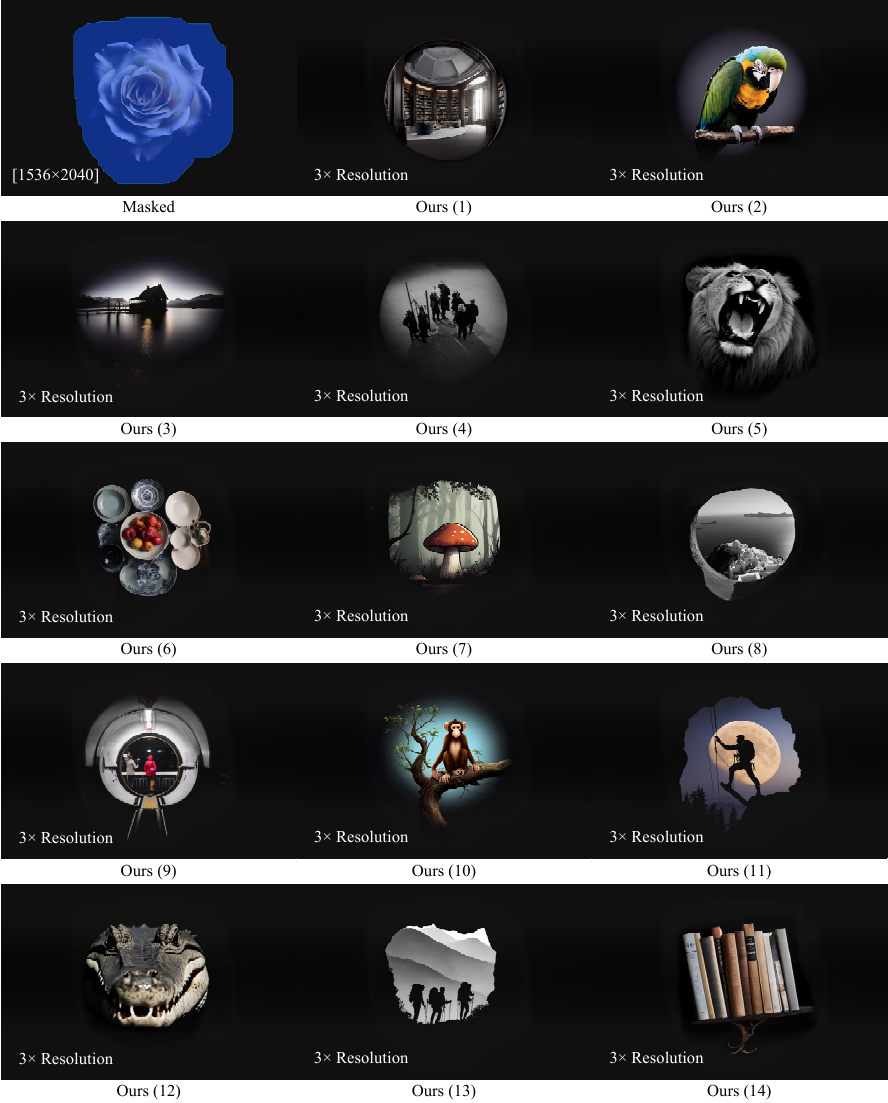}
    \caption{
    Visual results of UltraDiffEdit for high-resolution image editing using various text prompts.
    } 
    \label{fig:fig_supp_text_editing3}
\end{figure*}

\begin{figure*}[t]
    \centering
    \includegraphics[width=0.96\textwidth]{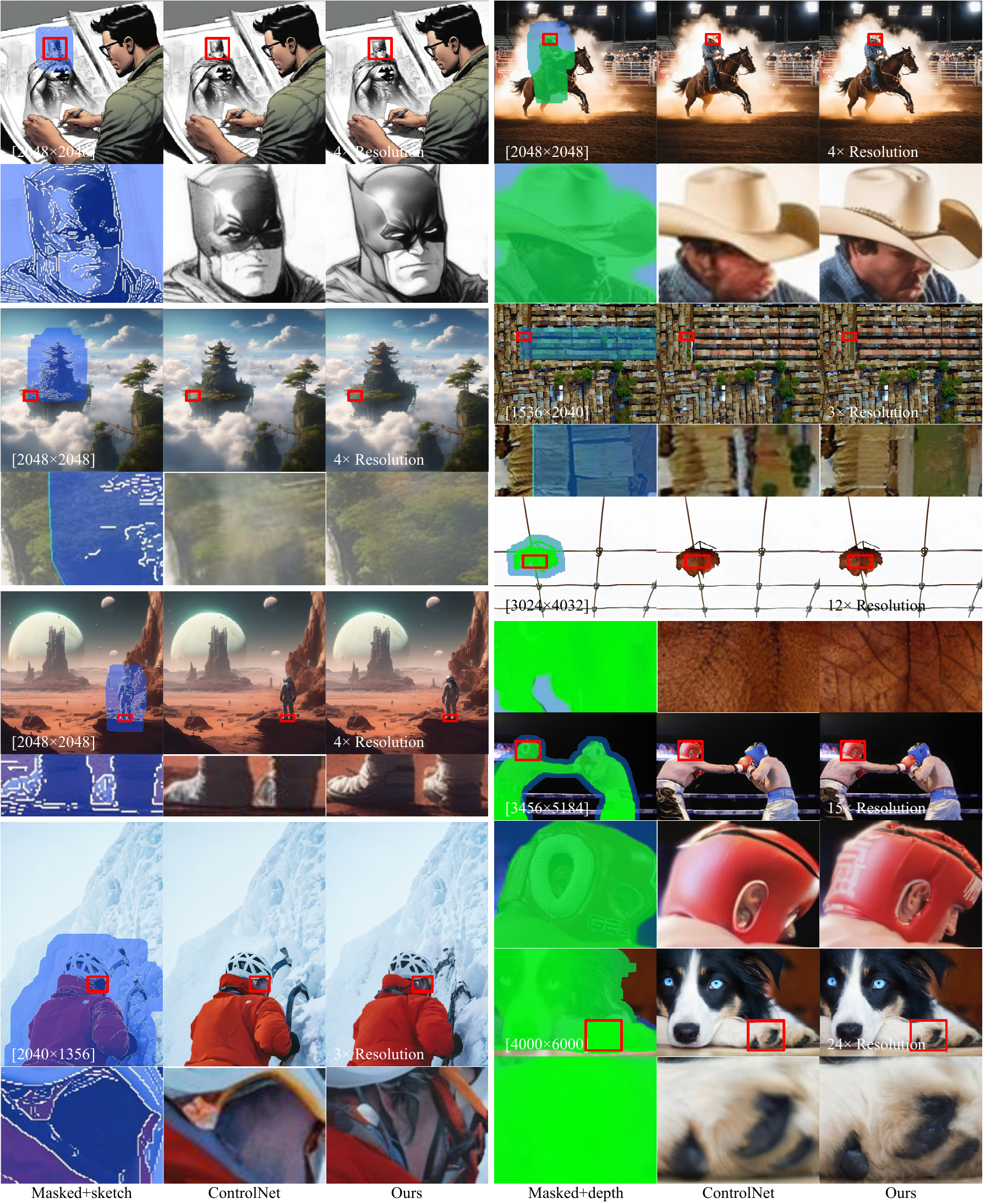}
    \caption{
    Visual results of extending ControlNet~\cite{zhang2023adding} to high-resolution image editing using Canny edges (left) and depth maps (right).
    } 
    \label{fig:fig_supp_sketch_depth_control}
\end{figure*}

\begin{figure}[t]
    \centering
    \includegraphics[width=0.48\textwidth]{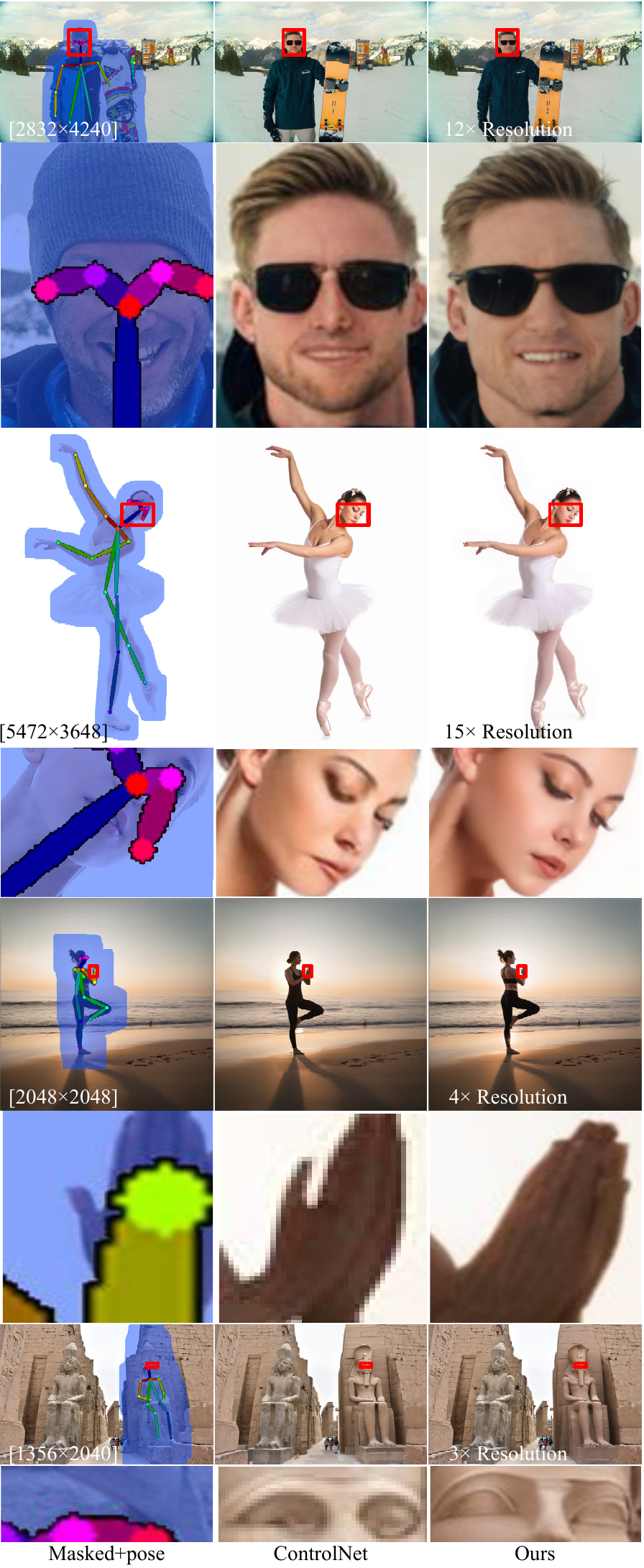}
    \caption{
    Visual results of extending ControlNet~\cite{zhang2023adding} to high-resolution image editing using pose keypoints.
    } 
    \label{fig:fig_supp_pose_control}
\end{figure}

\begin{figure}[t]
    \centering
    \includegraphics[width=0.48\textwidth]{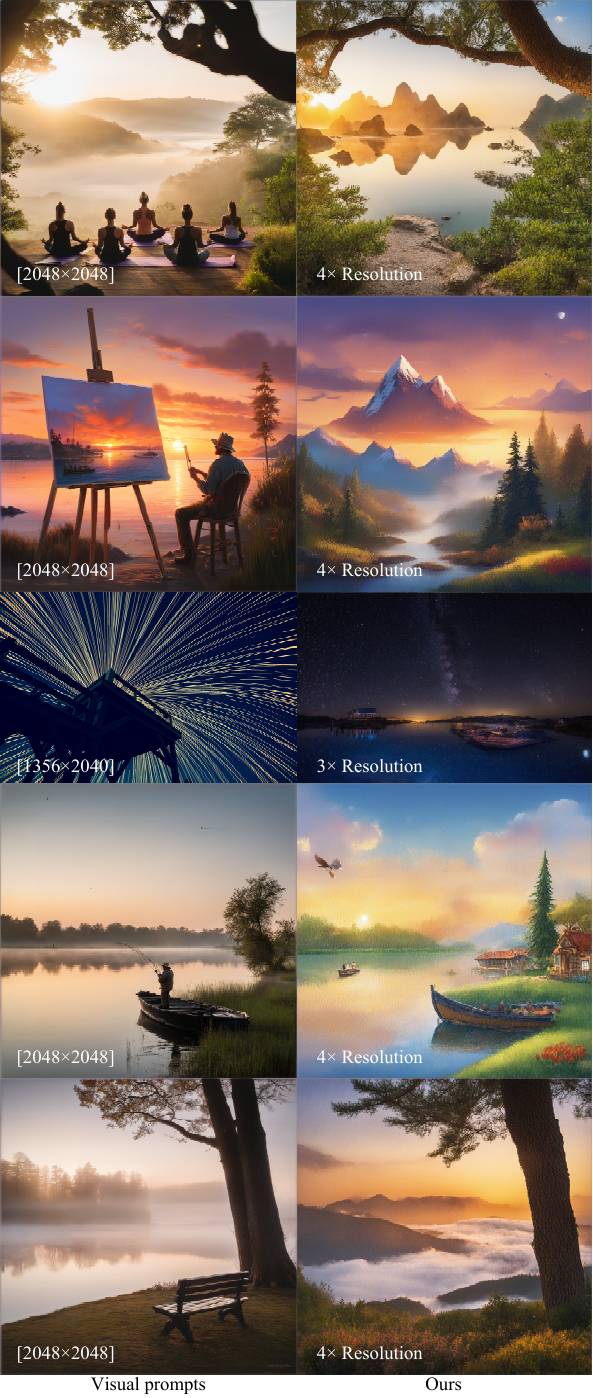}
    \caption{
    Visual results of extending IP-adapter~\cite{ye2023ip} to high-resolution image generation using visual prompts.
    } 
    \label{fig:fig_supp_IP}
\end{figure}

\section{More experimental results}\label{sec:exp_supp}

\textbf{More comparison results.}
Fig.~\ref{fig:fig_vis_supp_HURSD_compare1} and Fig.~\ref{fig:fig_vis_supp_HURSD_compare2} present additional results for ultra-high-resolution image editing, comparing our UltraDiffEdit with SDXL\cite{podell2023sdxl}+bicubic and DemoFusion~\cite{du2023demofusion}. 
While SDXL+bicubic and DemoFusion produce plausible results, SDXL+bicubic often generates blurred outputs, and DemoFusion can significantly alter background details, introduce distortions, or produce repetitive features, as noted in its original paper.
In contrast, our method preserves unedited regions while generating globally and locally consistent content. 
The global-local consistency denoising consistently integrates generated content into masked areas while maintaining the background. 
Additionally, our patch-based hybrid sampling captures local, intermediate, and global information effectively during the denoising process.

Fig.~\ref{fig:fig_vis_supp_HURSD_8K} provides more comparisons against DemoFusion~\cite{du2023demofusion} on 8K images. 
Fig.~\ref{fig:fig_supp_ultra} and Fig.~\ref{fig:fig_supp_ultra2} show more visual results for ultra-high-resolution image editing. Fig.~\ref{fig:fig_supp_text_editing}, Fig.~\ref{fig:fig_supp_text_editing2}, and Fig.~\ref{fig:fig_supp_text_editing3} show more high-resolution results using various text prompts.
More extension results for ControlNet using Canny edges, depth maps, and pose skeletons can be found in Fig.~\ref{fig:fig_supp_sketch_depth_control} and Fig.~\ref{fig:fig_supp_pose_control}; more generated results for IP-adapter are shown in Fig.~\ref{fig:fig_supp_IP}.
Note that the ``$x \times$ Resolution'' indicates the scaling factor relative to the original trained resolution.

\textbf{Prompts used in this paper.}
All prompts used in this paper are generated by BLIP-2~\cite{li2023blip} or ChatGPT~\cite{ChatGPT}. They are summarized here.

\noindent\textbf{Fig.~1 in the main text}:
\begin{itemize}
    \item A statue of a lion in front of a building.
\end{itemize}

\noindent\textbf{Fig.~2 in the main text}:
\begin{itemize}
    \item A picturesque lagoon with crystal clear waters, surrounded by cliffs and lush vegetation.
\end{itemize}

\noindent\textbf{Fig.~3 in the main text}:
\begin{itemize}
    \item A moonlit pond in a tranquil garden, water lilies floating gently, a perfect spot for nighttime reflection (top).
    \item A quiet cabin in the Canadian Rockies, smoke rising from the chimney into the crisp winter air (bottom).
\end{itemize}

\noindent\textbf{Figs.~4 and~5 in the main text}:
\begin{itemize}
    \item  A pair of worn hiking boots on a forest trail (row 1).
    \item  A futuristic highway, with autonomous vehicles zooming by under the glow of holographic billboards (row 2).
    \item  A comic book artist sketches a new superhero at a convention, fans watching eagerly (row 3).
\end{itemize}

\noindent\textbf{Fig.~6 in the main text}:
\begin{itemize}
    \item  The entrance to the temple of Karnak (top).
\end{itemize}

\noindent\textbf{Fig.~7 in the main text}:
\begin{itemize}
    \item A photo containing rock, telescope, binoculars, and flashlight.
    \item A photo containing treasure chest, and wizard.
    \item A detailed rendering of tent, and deer.
    \item A body of water
    \item A waterfall in the woods
\end{itemize}

\noindent\textbf{Fig.~8 in the main text}:
\begin{itemize}
    \item A blue vw beetle parked in the woods.
\end{itemize}

\noindent\textbf{Fig.~9 in the main text}:
\begin{itemize}
    \item  A wild Mustang herd gallops across the Nevada desert, a cloud of dust rising in their wake (top).
    \item  Ferns in the garden (bottom).
\end{itemize}

\noindent\textbf{Fig.~10 in the main text}:
\begin{itemize}
    \item  A blue tit.
\end{itemize}

\noindent\textbf{Fig.~11 in the main text}:
\begin{itemize}
    \item  A moth flying over a green background.
\end{itemize}

\noindent\textbf{Fig.~14 in the main text}:
\begin{itemize}
    \item  A majestic mountain range at dawn, the peaks illuminated by the first light of day.
\end{itemize}

\noindent\textbf{Fig.~15 in the main text}:
\begin{itemize}
    \item A mushroom on a log.
    \item A person climbing a mountain.
\end{itemize}

\noindent\textbf{Fig.~\ref{fig:fig_vis_supp_HURSD_compare1} in the supplement}:
\begin{itemize}
    \item  A dog with a long hair (top).
    \item  A dog walking on a street (bottom).

\end{itemize}

\noindent\textbf{Fig.~\ref{fig:fig_vis_supp_HURSD_compare2} in the supplement}:
\begin{itemize}
    \item  A fence with a dead leaf on it (top).
    \item  A dog running through the snow (bottom).
\end{itemize}

\noindent\textbf{Fig.~\ref{fig:fig_vis_supp_HURSD_8K} in the supplement}:
\begin{itemize}
    \item  A pink clematis flower on a brick wall (row 1).
    \item  A girl singing in a choir (row 2).
    \item  A girl reading a book (row 3).
    \item  A person walking down a wet street (row 4).
    \item  A young boy standing in front of a podium (row 5).

\end{itemize}

\noindent\textbf{Fig.~\ref{fig:fig_supp_ultra} in the supplement}:
\begin{itemize}
    \item A cupcake with white frosting (row 1).
    \item A drone flying over a field (row 2).
    \item A gazelle walking across a sandy area (row 3).
    \item A plate of food on a table (row 4).
\end{itemize}

\noindent\textbf{Fig.~\ref{fig:fig_supp_ultra2} in the supplement}:
\begin{itemize}
    \item A cat (row 1).
    \item A dog playing with a ball (row 2).
    \item Two chairs with covers on them (row 3).
    \item A dog with a long hair (row 4).
    \item A rock balancing on top of a rock (row 5).

\end{itemize}

\noindent\textbf{Fig.~\ref{fig:fig_supp_text_editing} in the supplement}:
\begin{itemize}
    \item A statue of St Peter holding a baby (row 1, ours (1)).
    \item A waterfall in the woods (row 1, ours (2)).
    \item A cheetah (row 2, ours (1)).
    \item A lion with his tongue out (row 2, ours (2)).
    \item The inside of an ancient building (row 3, ours (1)).
    \item A body of water (row 3, ours (2)).
    \item A statue (row 4, ours (1)).
    \item A penguin standing on a rock (row 4, ours (2)).
\end{itemize}

\noindent\textbf{Fig.~\ref{fig:fig_supp_text_editing2} in the supplement}:
\begin{itemize}

    \item A wolf looking to the side (case 1, ours (1)).
    \item A lion with his tongue out (case 1, ours (2)).

    \item A soldiers in camouflage (case 2, ours (1)).
    \item A penguin family (case 2, ours (2)).
    
    \item A woman wearing a hat (case 1, ours (1)).
    \item A mushroom in the woods (case 1, ours (2)).
    \item A statue (case 1, ours (3)).
    \item A dog with its mouth open (case 1, ours (4)).
    \item A flower with orange flowers (case 1, ours (5)).

\end{itemize}

\noindent\textbf{Fig.~\ref{fig:fig_supp_text_editing3} in the supplement}:
\begin{itemize}
    \item A library with a large circular room (ours (1)).
    \item A parrot sitting on a wooden post (ours (2)).
    \item A lake with a house on it (ours (3)).
    \item A group of people standing around a fence (ours (4)).
    \item A lion with his tongue out (ours (5)).
    \item A shelf with many different plates and bowls (ours (6)).
    \item A mushroom in the woods (ours (7)).
    \item The greek island of santorini (ours (8)).
    \item People standing on the platform of a train (ours (9)).
    \item A tree with a monkey tied to it (ours (10)).
    \item A person climbing a mountain (ours (11)).
    \item A crocodile in the sand (ours (12)).
    \item A group of people hiking (ours (13)).
    \item A bookshelf with many books on it (ours (14)).

\end{itemize}

\noindent\textbf{Fig.~\ref{fig:fig_supp_sketch_depth_control} in the supplement}:
\begin{itemize}
    \item A comic book artist sketches a new superhero at a convention, fans watching eagerly (left column, row 1).
    \item High above the clouds, a floating island hosts an ancient, forgotten temple (left column, row 2).
    \item A desolate alien planet, with a lone astronaut discovering the ruins of a long-lost civilization (left column, row 3).
    \item A person climbing a mountain (left column, row 4).
    \item A high-stakes rodeo competition in Texas, cowboys showcasing their skills under the bright arena lights (right column, row 1).
    \item A city with many houses and trees (right column, row 2).
    \item A fence with a dead leaf on it (right column, row 3).
    \item Two boxers in the ring (right column, row 4).
    \item A dog with blue eyes (right column, row 5).

\end{itemize}

\noindent\textbf{Fig.~\ref{fig:fig_supp_pose_control} in the supplement}:
\begin{itemize}
    \item A man holding a snowboard (row 1).
    \item A ballerina in a white tutue (row 2).
    \item An early riser practices yoga on a peaceful beach, the waves providing a calming soundtrack. (row 3).
    \item The entrance to the temple of Karnak. (row 4).
\end{itemize}

\bibliographystyle{IEEEtran}
\bibliography{main}

\vfill

\end{document}